\documentclass[journal,twoside]{IEEEtran}
\IEEEoverridecommandlockouts
\usepackage{cite}
\usepackage{amsmath,amssymb,amsfonts}
\usepackage{graphicx}
\usepackage{textcomp}
\usepackage{xcolor}

\usepackage[ruled]{algorithm2e}
\usepackage{subfigure}
\usepackage{bm}
\usepackage{amssymb}
\usepackage{threeparttable}
\usepackage{booktabs}
\usepackage{multirow}
\usepackage{color}

\def\BibTeX{{\rm B\kern-.05em{\sc i\kern-.025em b}\kern-.08em
    T\kern-.1667em\lower.7ex\hbox{E}\kern-.125emX}}
\begin{document}
	
\title{Memory Efficient Class-Incremental Learning for Image Classification}

\author{Hanbin Zhao, Hui Wang, Yongjian Fu, Fei Wu, Xi Li*
	
	\thanks{H. Zhao, H. Wang, Y. Fu and F. Wu are with College of Computer Science and Technology, Zhejiang University, Hangzhou 310027, China. (e-mail: {zhaohanbin, wanghui\_17, yjfu, wufei@zju.edu.cn})}%
	\thanks{X. Li* (corresponding author) is with the College of Computer Science and Technology, Zhejiang University, Hangzhou 310027, China and also with the Shanghai Institute for Advanced Study, Zhejiang University, Shanghai 201210, China (e-mail: {xilizju@zju.edu.cn} phone: 0571-87951247)}%
	\thanks{The first two authors contribute equally.}
}
\markboth{IEEE Transactions on Neural Networks and Learning Systems,~Vol.~XX, No.~X, 2021}%
{Zhao \MakeLowercase{\textit{et al.}}: Memory Efficient Class-Incremental Learning for Image Classification}
\maketitle

\begin{abstract}
With the memory-resource-limited constraints, class-incremental learning (CIL) 
usually suffers from the ``catastrophic forgetting'' problem when updating the 
joint classification model on the arrival of newly added classes. To cope with the forgetting problem, many CIL methods transfer the knowledge of old classes by preserving some exemplar samples into the size-constrained memory buffer. To utilize the memory buffer more efficiently, 
we propose to keep more auxiliary low-fidelity exemplar samples rather than the original real high-fidelity exemplar samples. Such memory-efficient exemplar preserving scheme make the old-class knowledge transfer more effective. 
However, the low-fidelity exemplar samples are often distributed in a different domain away from that of the original exemplar samples, that is, a domain shift. To alleviate this problem, we propose a duplet learning scheme that seeks to construct domain-compatible feature extractors and classifiers, which greatly narrows down the above domain gap. As a result, these low-fidelity auxiliary exemplar samples have the ability to moderately replace the original exemplar samples with a lower memory cost. In addition, we present a robust classifier adaptation scheme, which further refines the biased classifier (learned with the samples containing distillation label knowledge about old classes) with the help of the samples of pure true class labels. Experimental results demonstrate the effectiveness of this work against the state-of-the-art approaches.
\end{abstract}


\section{Introduction}\label{sec:introduction}
\IEEEPARstart{R}ecent years have witnessed a great development of incremental learning~\cite{xu2018reinforced, belouadah2018deesil, chaudhry2018riemannian,rannen2017encoder,aljundi2017expert, parisi2019continual,de2019continual, zhang2019class, hao2019end, hao2019take, lee2019overcoming, dong2018imbalanced,zhang2013load,gu2014incremental,penalver2012entropy,wang2019incremental,yang2019novel,nakamura2016nonparametric,chen2018universal},
which has a wide range of real-world applications with the capability of continual model learning.
To handle a sequential data stream with time-varying new classes,
class-incremental learning~\cite{rebuffi2017icarl} has emerged as a technique for the 
resource-constrained classification problem, which dynamically updates the model with the new-class
samples as well as a tiny portion of old-class information (stored in a limited memory buffer).
In general, class-incremental learning aims to set up a joint classification model simultaneously covering
the information from both new and old classes, and  
is usually facing the forgetting problem~\cite{french1999catastrophic, goodfellow2013empirical, mccloskey1989catastrophic, pfulb2019comprehensive, ritter2018online, liu2018rotate, serra2018overcoming, lee2017overcoming, krizhevsky2009learning,coop2013ensemble}
with the domination of new-class samples. To address the forgetting problem, many class-incremental learning
approaches typically concentrate on the following two aspects: 1) how to efficiently utilize the limited memory buffer (e.g., select representative exemplar samples from old classes); and 2) how to effectively attach the new-class samples with old-class information
(e.g., feature transferring from old classes to new classes or
distillation label~\cite{castro2018end} on each sample with old-class teacher model~\cite{li2018learning}). Therefore, we focus on effective class
knowledge transfer and robust classifier updating for class-incremental learning
within a limited memory buffer. 

\begin{figure}[t]
	\centering
	\includegraphics[width=1\columnwidth]{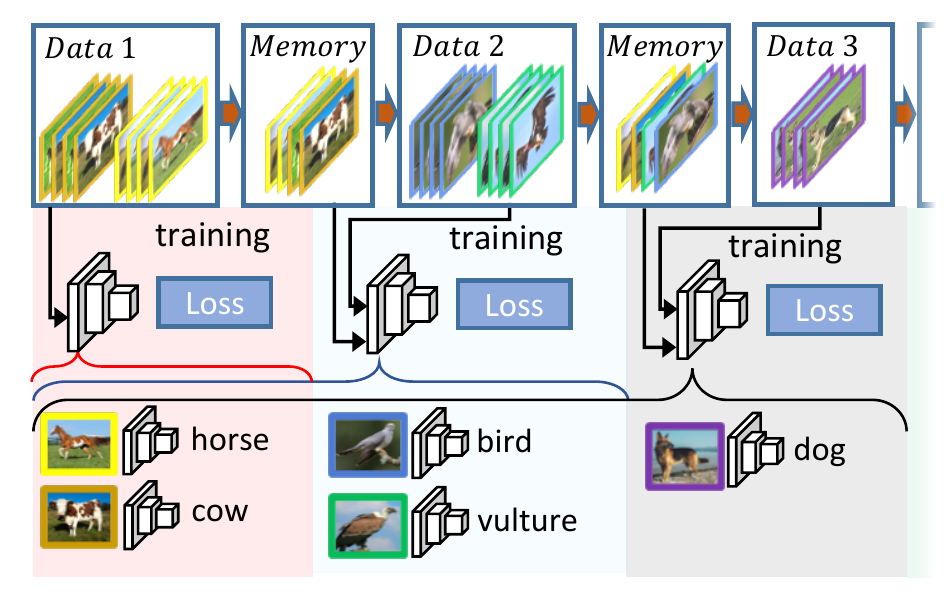}
	\caption{Illustration of resource-constrained class-incremental learning. Firstly a training is done on the first available data. After that, part of those data is stored in a limited memory. When new data arrives, the samples in the memory are extracted and used with the new data to train the network so that it can correctly identify all the classes it has seen.}
	\label{fig:class_incremental_learning}
\end{figure}
As for class knowledge transfer, a typical way is to preserve some exemplar samples into a memory buffer
which has a constrained size in practice. For maintaining a low memory cost of classification, existing approaches~\cite{castro2018end, rebuffi2017icarl}
usually resort to reducing the number of exemplar samples from old classes, resulting in the learning performance drop.
Motivated by this observation, we attempt to enhance the learning performance with a fixed memory buffer by 
increasing the number of exemplar samples while moderately reducing the fidelity of exemplar samples. Our goal is to build a memory efficient class-incremental learning manner with low-fidelity exemplars.
However, the normal exemplar-based class-incremental learning schemes~\cite{rebuffi2017icarl,castro2018end} can not work well with low-fidelity exemplars, because there exists a domain gap between the original exemplar samples and their corresponding
low-fidelity ones (with smaller memory sizes). 
Thus, a specific a learning scheme must be proposed to update the model while reducing the influence of domain shift.
In our duplet learning scheme, when facing the samples of new classes the low-fidelity exemplar samples
are treated as the auxiliary samples, resulting in a set of duplet sample pairs in the form of
original samples and their corresponding auxiliary samples. Based on such duplet sample pairs,
we construct a duplet-driven deep learner that aims to build domain-compatible feature extractors
and classifiers to alleviate the domain shift problem. With such a domain-compatible learning
scheme, the low-fidelity auxiliary samples have the capability of moderately replacing the
original high-fidelity samples, leading to more exemplar samples in the fixed memory buffer 
with a better learning performance.

After that, the duplet-driven deep learner
is carried out over the new-class samples to generate their corresponding distillation label information of old classes,
which makes the new-class samples inherit the knowledge of old classes.
In this way, the label information on each new-class sample is composed of both distillation labels
of old classes and true new-class labels.
Hence, the overall classifier is incrementally updated with these two kinds of label information.
Since the distillation label information is noisy, the classifer still has a small bias. Therefore, we propose a classifier adaptation scheme
to correct the classifier. Specifically, we fix the feature extractor learned with knowledge distillation, and then adapt the classifier over samples with true class labels only (without any
distillation label information). Finally, the corrected classifier is obtained as a more robust classifier. 

In summary, the main contributions of this work are three-fold. First, we propose a novel memory-efficient duplet-driven scheme for resource-constrained class-incremental learning, which innovatively utilizes low-fidelity auxiliary samples for old-class knowledge transfer instead of the original real samples. With more exemplar samples in the limited memory buffer, the proposed learning scheme is capable of learning domain-compatible feature extractors and classifiers, which greatly reduces the influence of the domain gap between the auxiliary data domain and the original data domain. Second, we present a classifier adaptation scheme, which refines the overall biased classifier (after distilling the old-class knowledge into the model) by using pure true class labels for the samples while keeping the feature extractors fixed. Third, extensive experiments over benchmark datasets demonstrate the effectiveness of this work against the state-of-the-art approaches.

The rest of the paper is organized as follows. We first describe the related work in Section \ref{related_work}, and then explain the details of our proposed strategy in Section \ref{method}.
In Section \ref{experiments}, we report the experiments that we conducted and discuss their results. Finally, we draw a conclusion and describe future work in Section \ref{conclusions}.

\section{Related Work}\label{related_work}
Recently, there have been a lot of research works on incremental learning with deep models~\cite{chen2017broad,xing2015perception,hou2017one,park2018incremental,sun2018concept,ding2019decode,liu2019flexible,lin2017end,guo2017zero, pan2019multiple, kuang2020causal,zhu2020dark,zhuang2017challenges, zhao2020few}. The works can be roughly divided into three fuzzy categories of the common incremental learning strategies.   

\subsection{Rehearsal strategies}
Rehearsal strategies~\cite{chaudhry2019agem,rebuffi2017icarl,castro2018end,belouadah2019il2m,xiang2019incremental,hou2019learning, Wu2019large} replay the past knowledge to the model periodically with a limited memory buffer, to strengthen connections for previously learned memories. Selecting and preserving some exemplar samples of past classes into the size-constrained memory is a strategy to keep the old-class knowledge.
A more challenging approach is pseudo-rehearsal with generative models. Some generative replay strategies~\cite{he2018exemplar,shin2017continual, lavda2018continual, wu2018incremental, van2018generative} attempt to keep the domain knowledge of old data with a generative model and using only generated samples does not give competitive results. ILCAN~\cite{xiang2019incremental} is one of the pseudo-rehearsal methods and takes the GANs conditioned on labeled embedding vectors as the generator and replays the past information. BiC~\cite{Wu2019large} adds a linear model after the last fully connected layer to correct bias towards new classes and ScaIL~\cite{belouadah2020scail} scales past classifiers' weights to make them more comparable to those of new classes. In contrast, our method focuses on class-incremental learning with low-fidelity exemplars and the biased classifier is refined by a classifier adaptation scheme.  

\subsection{Regularization strategies}
Regularization strategies extend the loss function with loss terms enabling the updated weights to retain past memories. The work in~\cite{li2018learning} preserves the model accuracy on old classes by encouraging the updated model to reproduce the scores of old classes for each image through knowledge distillation loss. 
The strategy in~\cite{shmelkov2017incremental} is to apply the knowledge distillation loss to incremental learning of object detectors.
Since the knowledge distillation loss often leads to noisy distillation label information of old classes on new-class samples, DECODE~\cite{ding2019decode} adopts a confidence module to enable label noise resistant classification.
Other strategies~\cite{kirkpatrick2017overcoming, zenke2017continual, schwarz2018progress, chaudhry2018riemannian} use a weighted quadratic regularization loss to penalize moving important weights used for old tasks. 

\subsection{Architectural strategies}
Architectural strategies~\cite{kemker2017fearnet, rusu2016progressive, lomonaco2017core50, maltoni2018continuous,mallya2018piggyback,rajasegaran2019random,li2019learn, huang2019neural,zhao2020and} keep the learned knowledge from the old classes and acquire the new knowledge from the new classes by manipulating the network architecture, e.g. parameter masking, network pruning. PNN~\cite{rusu2016progressive} combines the parameter freezing and network expansion, and CWR~\cite{lomonaco2017core50} is proposed with a fixed number of shared parameters based on PNN. LTG~\cite{li2019learn} proposed to expand the network architecture by neural architecture search (NAS) while retaining the previously learned knowledge.

\begin{figure*}[ht]
	\centering
	\subfigure{}{
		\begin{minipage}[ht]{1\columnwidth}
			\includegraphics[width = 1\columnwidth]{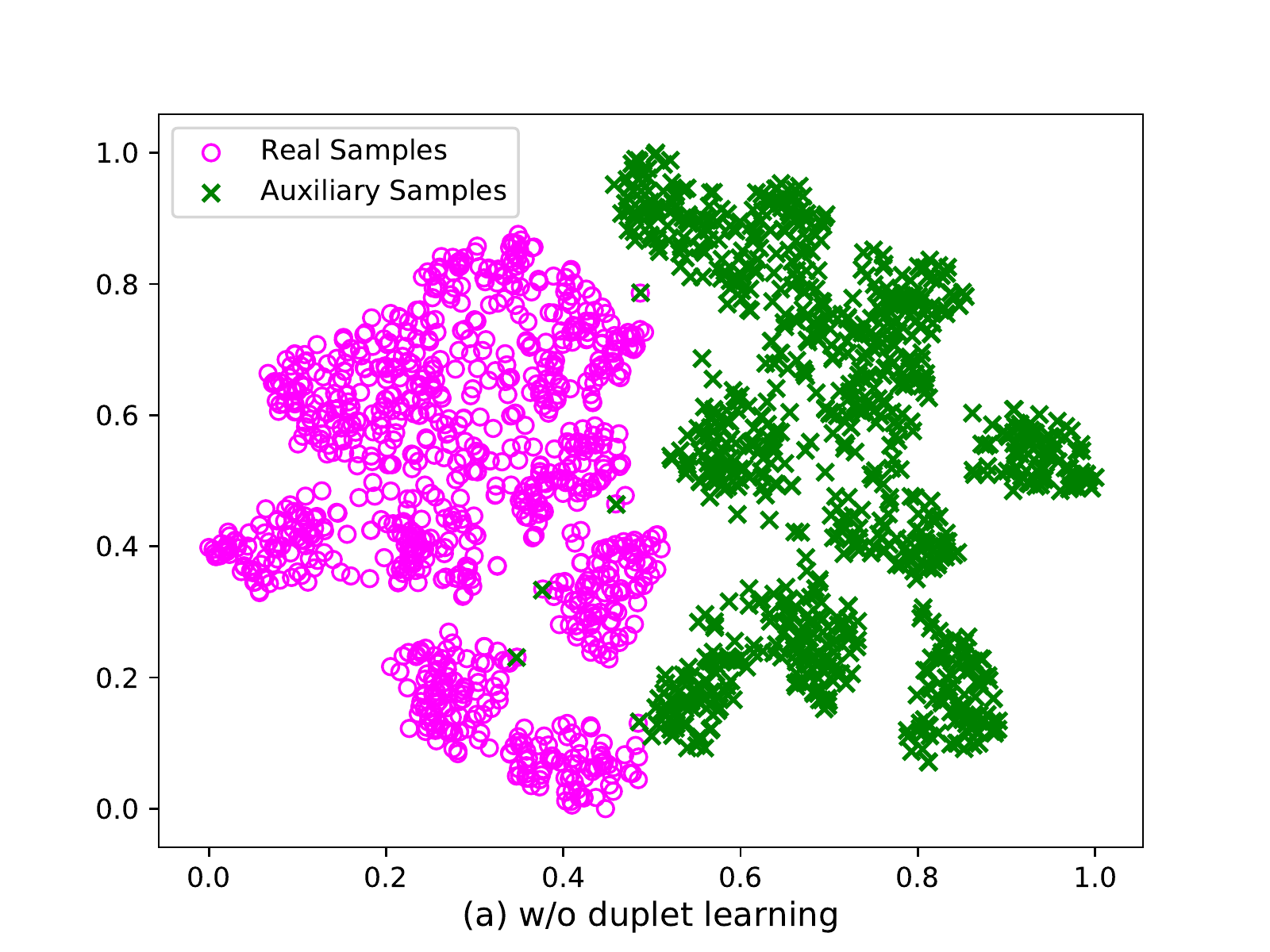}	
	\end{minipage}}
	\subfigure{}{
		\begin{minipage}[ht]{1\columnwidth}
			\includegraphics[width = 1\columnwidth]{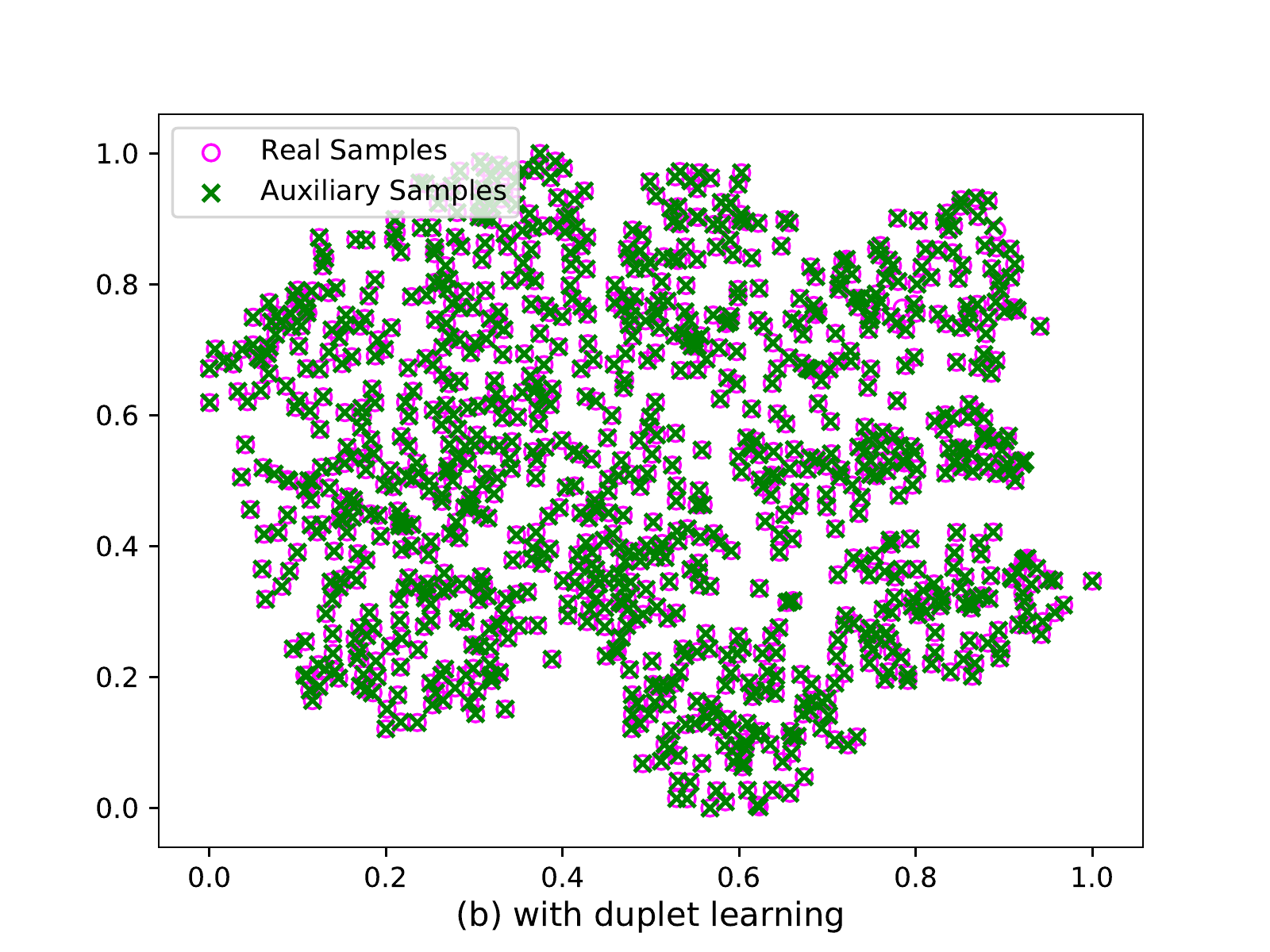}
		\end{minipage}	
	}
	\caption{Visualization of the low-fidelity auxiliary samples and the corresponding real samples with two different feature extractors by t-SNE. (a): The feature extractor is updated incrementally on the auxiliary data without our duplet learning scheme, and we notice that there is a large gap between the auxiliary data and real data; (b): The feature extractor is updated incrementally on the auxiliary data with our duplet learning scheme, and we observe that the domain gap is reduced. }
	\label{fig:tsne}
	\vspace{-10pt}	
\end{figure*}      

Our work belongs to the first and the second categories. We focus on effective past class knowledge transfer and robust classifier updating within a limited memory buffer. For effective class knowledge transfer with more exemplar samples, we innovatively design a memory-efficient exemplar preserving scheme and a duplet learning scheme that utilizes the low-fidelity exemplar samples for knowledge transfer, instead of directly utilizing the original real samples. Due to the domain gap between the low-fidelity exemplar samples and the corresponding original real samples, our method utilizes a duplet learning scheme to reduce it. Some of complex domain adaptation methods are proposed to reduce the domain gap between two different distributions~\cite{pan2010domain, tang2020unsupervised, xu2020reliable, li2018transfer, tzeng2017adversarial, courty2016optimal, niu2016action, ding2016incomplete}. Specifically, LPJT~\cite{li2019locality} focuses on joint feature adaptation and sample adaptation. HDAPA~\cite{li2018heterogeneous} is based on dictionary sharing coding to align the distribution gap on the new space. Besides, MDD~\cite{li2020maximum} pays attention to the loss design (i.e. maximum density divergence) to quantify the distribution divergence for adversarial training. Moreover, the distillation label information of old classes on new-class
samples with knowledge distillation is usually noisy. Motivated by this observation, we further refine the biased classifier in a classifier adaptation scheme.      

\begin{table}[ht]
	\centering
	\caption{Main notations and symbols used throughout the paper.}
	\resizebox{1\columnwidth}{!}{
		\Huge
		\begin{tabular}{c l l}
			\toprule[4pt]
			\textbf{Notation} & \multicolumn{2}{c}{\textbf{Definition}} \\
			\hline
			$X^h$&  \multicolumn{2}{l}{The sample set of class $h$}\\		
			$\widehat{X^h}$&  \multicolumn{2}{l}{Auxiliary form of the sample set of class $h$}\\	
			
			$A_t$&  \multicolumn{2}{l}{The added data of new classes at the $t$-th learning session}\\
			
			$F_t$&  \multicolumn{2}{l}{The deep image classification model at the $t$-th learning session}\\
			
			$\theta_{F_{t}}$&  \multicolumn{2}{l}{The parameters of $F_t$}\\
			
			$B_t$&  \multicolumn{2}{l}{The feature extractor of $F_t$}\\
			
			$C_t$&  \multicolumn{2}{l}{The classifier of $F_t$}\\
			
			$\widehat{P^j_{t}}$&  \multicolumn{2}{l}{The auxiliary exemplar samples of old class $j$ stored at the $t$-th learning session}\\
			
			$\widehat{R_{t}}$&  \multicolumn{2}{l}{The auxiliary exemplar samples of old classes preserved at previous $t$ learning sessions}\\
			
			$E$&  \multicolumn{2}{l}{The mapping function of the encoder}\\
			
			$D$&  \multicolumn{2}{l}{The mapping function of the decoder}\\
			\hline		
			\bottomrule[4pt]		
		\end{tabular}%
	}
	
	\label{Notation}%
\end{table}%

\section{Method}\label{method}
\subsection{Problem Definition}\label{problem_definition}
Before presenting our method, we first provide an illustration of the main notations and symbols used hereinafter (as shown in Table~\ref{Notation}) for a better understanding. 

Class-incremental learning assumes that samples from one new class or a batch of new classes come at a time. For simplicity, we suppose that the sample sets in a data stream arrive in order (i.e. $X^1, X^2, \dots$), and the sample set $X^h$ contains the samples of class $h$ ($h\in\{1, 2, \dots\}$). We consider the time interval from the arrival of the current batch of classes to the arrival of the next batch of classes as a class-incremental learning session~\cite{he2018exemplar, kemker2017fearnet}. A batch of new class data added at the $t$-th ($t\in\{1, 2, \dots\}$) learning session are represented as:
\begin{equation}
A_{t} = 
\begin{cases}
X^{1} \cup \dots \cup X^{k} \cup \dots \cup X^{n_{1}} & t=1\\
X^{n_{t-1}+1} \cup \dots \cup X^{k} \cup \dots \cup X^{n_{t}} & t \geq 2
\end{cases}
\label{equation_two}
\end{equation}

In an incremental learning environment with a limited memory buffer, previous samples of old classes can not be stored entirely and only a small number of exemplars from the samples are selected and preserved into memory for old-class knowledge transfer~\cite{castro2018end, rebuffi2017icarl}. The memory buffer is dynamically updated at each learning session.

At the $t$-th session, after obtaining the new-class samples $A_t$ we access memory $M_{t-1}$ to extract the exemplar samples of old-class information. Let $P^j_{t-1}$ denote the set of exemplar samples extracted from the memory at the $t$-th session for the old class $j$:
\begin{equation}
P^j_{t-1} = \left\{(x^j_1, y^j), \dots, (x^j_{e_{t-1}}, y^j)\right\}
\end{equation}
where $e_{t-1}$ is the number of exemplar samples and $y^j$ is the corresponding ground truth label. $P^j_{t-1}$ is the first $e_{t-1}$ samples selected from the sorted list of samples of class $j$ by herding~\cite{welling2009herding}. 
And then $M_{t-1}$ can be rewritten as:
\begin{equation}
M_{t-1} = P^1_{t-1} \cup \dots \cup P^{n_{t-1}}_{t-1}
\end{equation}

The objective is to train a new model $F_t$ which has competitive classification performance on the test set of all the seen classes. $F_t$ represents the deep image classification model at the $t$-th learning session and the parameters of the model are denoted as $\theta_{F_t}$. The output of $F_t$ is defined as:
\begin{equation}
F_{t}(x) = [F^1_{t}(x),\dots,F^{n_{t-1}}_{t}(x),F^{n_{t-1}+1}_{t}(x),\dots,F^{n_{t}}_{t}(x)]
\end{equation} 
$F_t$ is usually composed of a feature extractor $B_t$ and a classifier $C_t$. After obtaining the model $F_t$, the memory buffer is updated and $M_{t}$ is constructed with the exemplars in $A_t$ and a subset of $M_{t-1}$. 

In the experiments, we use a herding selection strategy (as utilized in iCaRL [20], not a random selection strategy) to select new-class exemplars to write in memory and remove some of the stored old-class exemplars when the memory is used up. In principle, the herding selection strategy [71] corresponds to the priority sorting which ranks the instances of each individual class separately according to their distances to class center in the ascending order. As a result, we have a set of class-specific ranking lists, each of which is associated with a sequence of instances. As for the selection of new-class exemplars, we pick out the top $K$ instances of the ranking list as the exemplars for each new class. As for the removal of old-class instances, the bottom ones of each old-class ranking list are accordingly removed.

\begin{figure*}[t]
	\centering
	\includegraphics[width=1\textwidth]{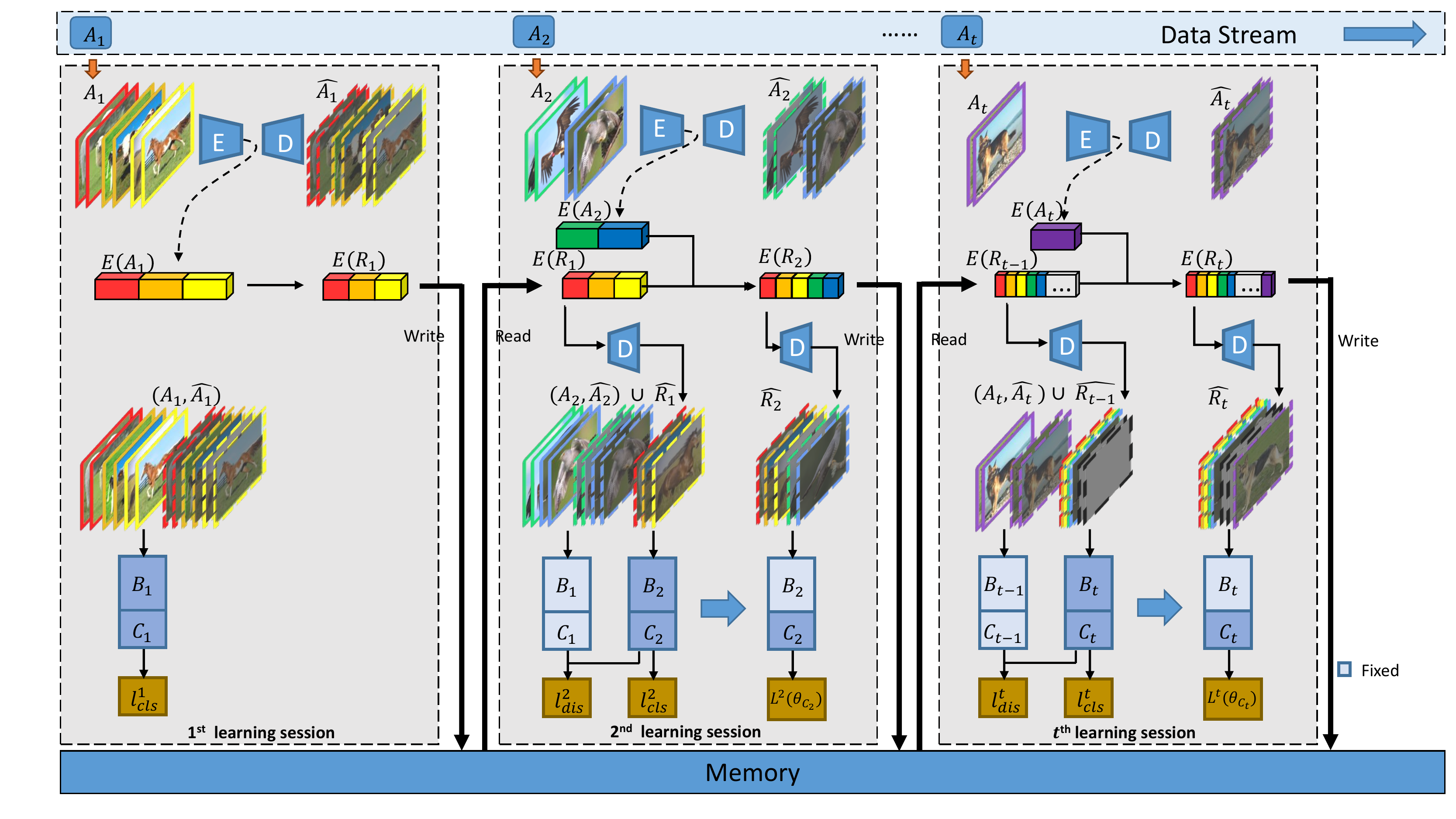}
	\caption{Illustration of the process of class-incremental learing with our duplet learning scheme and classifier adaptation scheme. The notations of ``fixed", ``write", and ``read" means fixing the parameters, storing the auxiliary exemplars in the memory, and obtaining the auxiliary exemplars from the memory respectively.
	We use $B_t$ and $C_t$ ($t\in\mathbb{N}$) to represent the feature extractor and the classifier respectively at the $t$-th learning session. For initialization, $F_1$ is trained from scratch with the set of duplet sample pairs $(A_1, \widehat{A_1})$, $\widehat{R_1}$ is then constructed and stored in the memory in the form of $E(R_1)$. At the $t$-th learning session, firstly we train a new feature extractor $B_t$ and a biased classifier on all the seen classes with our duplet learning scheme, and the exemplar auxiliary samples $\widehat{R_t}$ for all the seen classes are then constructed. Finally, we update the classifier with the classifier adaptation scheme on $\widehat{R_t}$.}
	\label{fig:our_scheme}
	\vspace{-10pt}	
\end{figure*}
\subsection{Memory Efficient Class-incremental Learning}\label{our_auxiliary_samples}
It is helpful for knowledge transfer to store more number of exemplars in the memory~\cite{castro2018end}.
For effective knowledge transfer, we propose a memory efficient class-incremental learning manner, which means utilizing more low-fidelity auxiliary exemplar samples to approximately replace the original real exemplar samples. 

We present an encoder-decoder structure to transform the original high-fidelity real sample $x$ to the corresponding low-fidelity sample $\widehat{x}$ (with smaller memory size):
\begin{equation}
\widehat{x} = D \circ E(x)
\label{decoder_encoder}
\end{equation}
where $E$ is the mapping function of the encoder and $D$ is the mapping function of the decoder. 
Due to the loss of fideltiy, we keep the auxiliary sample code with smaller memory cost compared with storing the corresponding real sample. 
We use $r$ to represent the cost ratio of keeping a low-fidelity sample and keeping the corresponding high-fidelity real sample ($r = \frac{size(E(x))}{size(x)}$). This cost ratio corresponds to the compression cost ratio, which is only used as the measure of calculating the compression cost ratio of one sample after compression to that before compression.

Let $\widehat{P^j_{t-1}}$ denote the set of exemplar auxiliary samples extracted from the memory at the $t$-th session for the old class $j$:
\begin{equation}
\widehat{P^j_{t-1}} = \left\{(\widehat{x^j_1}, y^j), \dots, (\widehat{x^j_{\widehat{e_{t-1}}}}, y^j)\right\}
\end{equation}
where $\widehat{e_{t-1}}$ is the number of exemplar auxiliary samples. With the fixed size memory buffer, $\widehat{e_{t-1}}$ is larger than $e_{t-1}$.
We use $\widehat{R_{t-1}}$ to represent all the exemplar auxiliary samples extracted from $M_{t-1}$:
\begin{equation}
\widehat{R_{t-1}} = \widehat{P^1_{t-1}} \cup \dots \cup \widehat{P^{n_{t-1}}_{t-1}}
\end{equation}
And then $M_{t-1}$ in our memory efficient class-incremental learning can be represented as:
\begin{equation}
M_{t-1} = E(R_{t-1})
\end{equation}
where $R_{t-1}$ represents the corresponding real samples of $\widehat{R_{t-1}}$. 

At the $t$-th session, we use the exemplars $\widehat{R_{t-1}}$ and newly added data $A_t$ to train the model $F_t$. More number of exemplars from $M_{t-1}$ are helpful for learning the model. However, the domain gap between the auxiliary samples and their original versions (as shown in Figure~\ref{fig:tsne}(a)) often gives a bad performance. In order to fix this issue, we propose a duplet class-incremental learning scheme in the following subsection.

\subsection{Duplet Class-incremental Learning Scheme}\label{duplet_training}
To reduce the influence of domain shift, we propose a duplet learning scheme, which trains the model using a set of duplet sample pairs in the form of original samples and their corresponding auxiliary samples. 

A duplet sample pair $(x, \widehat{x}, y)$ is constructed from an auxiliary sample $(\widehat{x},y)$ and the corresponding real sample $(x,y)$. At the $t$-th learning session (as shown in Figure~\ref{fig:our_scheme}), we construct a set of duplet sample pairs with $A_t$ and $\widehat{A_t}$, denoted as:
\begin{equation}
\begin{array}{ll}
(A_t,\widehat{A_t}) = \left\{(x_k, \widehat{x_k},y_k)\right\}^{\left|A_t\right|}_{k=1} \\
s.t.~(x_k,y_k) \in A_t, (\widehat{x_k}, y_k) \in \widehat{A_t}
\end{array}
\end{equation}  
We train the model $F_t$ with the duplet sample pairs of $(A_t,\widehat{A_t})$ and the exemplar auxiliary samples of old classes $\widehat{R_{t-1}}$ by optimizing the objective function $L^t$:
\begin{equation}
\begin{array}{ll}
L^t(\theta_{F_t}) = L^t_{1}(\theta_{F_t}) + L^t_{2-dup}(\theta_{F_t})
\end{array}
\label{L_t_DUP}
\end{equation}  
where $L^t_{1}$ is the loss term for $\widehat{R_{t-1}}$ and $L^t_{2-dup}$ is the loss term for $(A_t,\widehat{A_t})$.

\noindent{$\bm{L^t_{1}}$\textbf{:}} For preserving the model performance on old-class auxiliary samples, $L^t_{1}$is defined as:
\begin{equation}
\begin{array}{ll}
L^t_{1}(\theta_{F_t}) = \frac{1}{\left|\widehat{R_{t-1}}\right|}{\sum\limits_{(\widehat{x},y) \in \widehat{R_{t-1}}}} l^t((\widehat{x},y); \theta_{F_t})  
\end{array}
\end{equation} 
where $l^t$ is composed of a classification loss term $l^t_{cls}$ and a knowledge distillation loss term $l^t_{dis}$ for one sample, which is defined as:
\begin{equation}
\begin{array}{ll}
l^t((x,y); \theta_{F_t}) = l^t_{cls}((x,y); \theta_{F_t}) + \lambda l^t_{dis}((x,y); \theta_{F_t})
\end{array}
\end{equation}
where $\lambda$ is a balance scalar to make a trade-off between $l^t_{cls}$ and $l^t_{dis}$ and we use $\lambda=1.0$ as default in our experiments.

The classification loss $l^t_{cls}$ for one training sample on newly added classes is formulated as:         
\begin{equation}
\begin{array}{ll}
l^t_{cls}((x,y); \theta_{F_t}) = \sum\limits_{k=n_{t-1}+1}^{n_t} Entropy(F^k_t(x;\theta_{F_t}),\delta_{y=k})
\end{array}
\end{equation}
where $\delta_{y=k}$ is a indicator function and denoted as:
\begin{equation}
\begin{array}{ll}
\delta_{y=k} =
\begin{cases}
1  & y=k\\
0  & y \neq k
\end{cases}
\end{array}
\end{equation}
$Entropy(\cdot,\cdot)$ is a cross entropy function and represented as:
\begin{equation}
Entropy(\widehat{y},y) = -[y\log(\widehat{y})+(1-y)\log(1-\widehat{y})]
\end{equation} 

The knowledge distillation loss $l^t_{dis}$ is defined as:
\begin{equation}
\begin{array}{ll}
l^t_{dis}((x,y); \theta_{F_t}) = \sum\limits_{k=1}^{n_{t-1}} Entropy(F^k_t(x;\theta_{F_t}),F^k_{t-1}(x;\theta_{F_{t-1}}))
\end{array}
\label{l_dis}
\end{equation} 
its aim is to make the output of the model $F_t$ close to the distillation class label $F^k_{t-1}(x;\theta_{F_{t-1}})$ ($k\in \left\{1,2,\dots,n_{t-1}\right\}$) of the previously learned model $F_{t-1}$ on old classes.

\noindent{$\bm{L^t_{2-dup}}$ \textbf{:}} For the new-class duplet sample pairs, $L^t_{2-dup}$ encourages the output of the model on the real sample similar to that on the corresponding auxiliary sample and is defined as:
\begin{equation}
\begin{array}{ll}
L^t_{2-dup}(\theta_{F_t}) = \frac{1}{\left|A_{t}\right|+\left|\widehat{A_{t}}\right|}{\sum\limits_{(x,\widehat{x},y) \in  (A_{t},\widehat{A_{t}})} } [l^t((x,y);\theta_{F_t})
\\
+l^t((\widehat{x},y);\theta_{F_t})] 
\end{array}
\end{equation}

In general, we can obtain a domain-compatible feature extractor and a classifier on all the seen classes by optimizing the loss function in Equation~\eqref{L_t_DUP}. The domain gap in Figure~\ref{fig:tsne}(a) is decreased greatly with our duplet learning scheme as shown in Figure~\ref{fig:tsne}(b).

\subsection{Classifier Adaptation}\label{coarse_to_fine}
Our duplet-driven deep learner is carried out over the new classes samples to generate their corresponding distillation label information of old classes through the distillation loss (as defined in Equation~\eqref{l_dis}), which makes the new-class samples inherit the knowledge of old classes. 
Since the distillation label knowledge is noisy, we propose a classifier adaptation scheme to refine the classifier over samples with true class labels only (without any distillation label information).

Taking the $t$-th learning session for example, we can obtain a domain-compatible but biased classifier $C_t$ by optimizing the objective function defined in Equation~\eqref{L_t_DUP}. 
Here, we fix the feature extractor $B_t$ learned and continue to optimize the parameters of the classifier further only using the true class label knowledge. Then, the optimization is done by using exclusively the auxiliary samples that are going to be stored in memory $M_t$. The objective function is formulated as below:
\begin{equation}
\small
\begin{array}{ll}
L^t(\theta_{C_t}) = \frac{1}{\left|\widehat{R_{t}}\right|}{\sum\limits_{(\widehat{x},y) \in \widehat{R_{t}}}}\sum\limits_{k=1}^{n_t} Entropy(C^k_t \circ B_t(\widehat{x}; \theta_{C_t})), \delta_{y=k})
\end{array}
\label{L_t_CTF}
\end{equation}

By minimizing the objective function, the classifier $C_t$ is refined and has better performance for all the seen classes.
\begin{algorithm}[t]
	\caption{Training the model at the $t$-th session}
	\label{algorithm_Updating}
	\KwIn{The added data of new classes $A^t$}
	\textbf{Require:} The exemplar auxiliary samples $\widehat{R^{t-1}}$ and the parameters $\theta_{F_{t-1}} = \left\{\theta_{C_{t-1}}, \theta_{B_{t-1}}\right\}$ \\
	\everypar={\nl}
	Obtain the auxiliary samples $\widehat{A^t}$ for new classes from $A^t$ using Equation~\eqref{decoder_encoder}\;
	Initialize $\theta_{B_t}$, $\theta_{C_t}$ with $\theta_{B_{t-1}}$ and $\theta_{C_{t-1}}$ respectively\;
	\tcc{The duplet learning scheme}
	\everypar={\nl}
	Obtain the optimal parameters $\theta_{B_t}$ of the feature extractor and a biased classifier by minimizing Equation~\eqref{L_t_DUP}\;
	Obtain the exemplar auxiliary samples $\widehat{R^{t}}$ of all the seen classes from $\widehat{R^{t-1}}$ and $\widehat{A^t}$ (described in Section \ref{our_auxiliary_samples})\;	
	\tcc{The classifier adaptation scheme}
	\everypar={\nl}
	Fix the parameters $\theta_{B_t}$ and adapt the biased classifier by minimizing Equation~\eqref{L_t_CTF} to obtain the optimal parameters $\theta_{C_t}$\;
	\KwOut{The auxiliary exemplar samples $\widehat{R^{t}}$ and the parameters $\theta_{F_{t}} = \left\{\theta_{C_{t}}, \theta_{B_{t}}\right\}$}
\end{algorithm}
Figure~\ref{fig:our_scheme} illustrates the process of class-incremental learning with our duplet learning scheme and classifier adaptation in detail.
For initialization, $F_1$ is trained from scratch with the set of duplet sample pairs $(A_1, \widehat{A_1})$, then $\widehat{R_{1}}$ is constructed and stored in the memory in the form of $E(R_{1})$. At the $t$-th learning session, we extract the auxiliary samples $\widehat{R_{t-1}}$ for previous classes and $\widehat{A_t}$ for new classes. Firstly we train a domain-compatible feature extractor $B_t$ and a classifier on all the seen classes with our duplet learning scheme (described in Section~\ref{duplet_training}). Then the exemplar auxiliary samples $\widehat{R_{t}}$ for all the seen classes are constructed from $\widehat{R_{t-1}}$ and $\widehat{A_t}$ (described in Section~\ref{our_auxiliary_samples}). Finally we update the classifier further with our classifier adaptation scheme (introduced in Section~\ref{coarse_to_fine}).  Algorithm~\ref{algorithm_Updating} lists the steps for training the model in detail.

\section{Experiments}\label{experiments} 
\subsection{Datasets}
\noindent\textbf{CIFAR-100}~\cite{krizhevsky2009learning} is a labeled subset of the 80 million tiny images dataset for object recognition. This dataset contains 60000 $32\times32$ RGB images in 100 classes, with 500 images per class for training and 100 images per class for testing.  

\noindent\textbf{ILSVRC}~\cite{krizhevsky2012imagenet} is a dataset for ImageNet Large Scale Visual Recognition Challenge 2012. It contains 1.28 million training images and 50k validation images in 1000 classes.

\subsection{Evaluation Protocol}
We evaluate our method on the iCIFAR-100 benchmark and the iILSVRC benchmark proposed in~\cite{rebuffi2017icarl}. On iCIFAR-100, in order to simulate a class-incremental learning process, we train all 100 classes in batches of 5, 10, 20, or 50 classes at a time, which means 5, 10, 20, or 50 classes of new data are added at each learning session. The order of added classes are arranged in a fixed random order. After each batch of classes are added, the obtained accuracy is computed on a subset of test data sets containing only those classes that have been added. The results we report are the average accuracy without considering the accuracy of the first learning session as it does not represent the incremental learning described in~\cite{castro2018end}. On iILSVRC, we use a subset of 100 classes which are trained in batches of 10 (iILSVRC-small) and the data of 100 classes are randomly sampled from ILSVRC~\cite{rebuffi2017icarl}. For a fair comparison, we take the same experimental setup as that of~\cite{rebuffi2017icarl}, which randomly selects 50 samples for each class of ILSVRC as the test set and the rest as the training set. In addition, the evaluation method of the result and the order arrangement of the added classes are the same as that on the iCIFAR-100. The redundancy among training samples mainly comes from two aspects: one is from the sample space containing the inter-sample content redundancy between samples, and the other is associated with the fidelity of a sample with different image qualities. Such two aspects can be respectively reflected by two kinds of ratio measures: the inter-sample content cost ratio and the compression cost ratio. For better clarity, we utilize the memory cost ratio to represent the total cost ratio of few low-fidelity representative old-class samples to all real old-class samples in class-incremental learning. Mathematically, the memory cost ratio is defined as the product of the inter-sample content cost ratio and the compression cost ratio. Specifically, our method firstly reduces inter-sample content redundancy by selecting several representative samples per class and keep these exemplars in a memory buffer for old-class knowledge preserving (associated with inter-sample content cost ratio), and then reduces intra-sample content redundancy by compressing each sample (corresponding to compression cost ratio). For example, if our method selects 80 samples per class on iCIFAR-100 and compresses each sample with compression cost ratio being $\frac{1}{4}$, the memory cost ratio is equal to $\frac{80}{500} \times \frac{1}{4} = 4.00\%$, which is practically reasonable for real class-incremental learning applications.

\begin{figure}[t]
	\centering
	\includegraphics[width=0.48\textwidth]{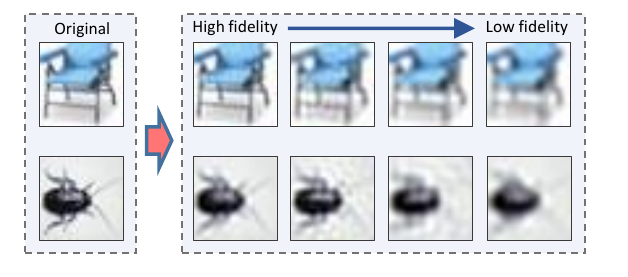}
	\caption{Illustration of original real samples from CIFAR-100 and their auxiliary sample equivalents with different fidelities.}
	\label{fig:low_fidelity_samples}
\end{figure}

\subsection{Implementation Details}
\paragraph{Data Preprocessing}\label{preprocessing} on CIFAR-100, the only preprocessing we do is the same as iCaRL~\cite{rebuffi2017icarl}, including random cropping, data shuffling and per-pixel mean subtraction. On ILSVRC, the augmentation strategy we use is the $224 \times 224$ random cropping and the horizontal flip.

\paragraph{Generating auxiliary samples of different fidelities} we generate auxiliary samples of different fidelities (as shown in Figure~\ref{fig:low_fidelity_samples}) with various kinds of encoder-decoder structures (e.g. PCA, Downsampling and Upsampling). The fidelity factors $r$ we use (as defined in Section 3.1) are as follow: for iCIFAR-100, we use $\frac{1}{3}$, $\frac{1}{6}$ and $\frac{1}{12}$ when using a PCA based reduction and $\frac{1}{3}$, $\frac{1}{6}$ when using downsampling. For iILSVRC, we conduct experiments with $r$ values of $\frac{1}{4}$ based on downsampling. We use PCA based method to construct the encoder-decoder structure on mini-batch level. In the experiments, the decoder is updated dynamically by incremental-PCA~\cite{artac2002incremental} as the new session arrives. Hence, the decoder is changing in different stream. In our experiments, we use a memory buffer of $2000$ full samples. Since the sample fidelity and number of exemplars is negatively correlated, our approach can store $\frac{2000}{r}$ samples. 
\begin{table}[t]
	\centering
	\caption{Evaluation of different methods which either using the auxiliary exemplars or real exemplars. Added classes at each session is $10$.}
	\resizebox{0.8\columnwidth}{!}{
		\begin{threeparttable}
			\begin{tabular}{llc}
				\toprule
				Method&Exemplar&Average Accuracy\\
				\midrule
				\midrule
				\multirow{2}{*}{iCaRL}&Real& \textbf{60.79$\pm$0.34\%}\\
				~&Auxiliary& 50.86$\pm$0.47\%\\
				\midrule
				\multirow{2}{*}{iCaRL-Hybrid1}&Real& \textbf{55.10$\pm$0.55\%}\\
				~&Auxiliary& 44.86$\pm$0.32\%\\
				\midrule
				\multirow{2}{*}{Ours.FC}&Real&61.67$\pm$0.18\%\\
				~&Auxiliary&\textbf{67.04$\pm$0.21\%}\\
				\midrule
				\multirow{2}{*}{Ours.NCM}&Real&61.97$\pm$0.10\%\\
				~&Auxiliary&\textbf{66.95$\pm$0.12\%}\\
				\bottomrule
			\end{tabular}
		\end{threeparttable}
	}
	
	\label{tab:ablation_experiments_exemplar}%
\end{table}%
\paragraph{Training details} for iCIFAR-100 at each learning session, we train a 32-layers ResNet~\cite{he2016deep} using SGD with a mini-batch size of $256$ ($128$ duplet sample pairs are composed of $128$ original samples and $128$ corresponding auxiliary samples) by the duplet learning scheme.  The initial learning rate is set to $2.0$ and is divided by $5$ after $49$ and $63$ epochs. We train the network using a weight decay of $0.00001$ and a momentum of $0.9$. For the classifier adaptation, we use the auxiliary exemplar samples for normal training and the other parameters of the experiments remain the same. We implement our framework with the theano package and use an NVIDIA TITAN 1080 Ti GPU to train the network. For iILSVRC, we train an 18-layers ResNet~\cite{he2016deep} with the initial learning rate of $2$, divided by $5$ after $20$, $30$, $40$ and $50$ epochs. The rest of the settings are the same as those on the iCIFAR-100. In the experiments, we did not carefully tune $\lambda$ and directly used $\lambda=1.0$ as the default value for all the cases. For epoch information, 70 epochs are utilized at each incremental learning session on iCIFAR-100 (60 epochs on iILSVRC-small and 100 epochs on iILSVRC-full). For the repeated times of experiments, we repeated each experiment for 5 times.

\subsection{Ablation Experiments}
In this section, we first evaluate different methods which either using the auxiliary exemplars or real exemplars. Then we carry out two ablation experiments to validate our duplet learning and classifier adaptation scheme on iCIFAR-100, and we also conduct another ablation experiment to show the effect of the auxiliary sample's fidelity and the auxiliary exemplar data size for each class when updating the model. Finally, we evaluate methods with memory buffer of different sizes.

\subsubsection{Baseline}
for the class-incremental learning problem, we consider three kinds of baselines, which are: a) LWF.MC~\cite{li2018learning}, utilizes knowledge distillation in the incremental learning problem, b) iCaRL~\cite{rebuffi2017icarl}, utilizes exemplars firstly for old-class knowledge transfer and a nearest-mean-of-exemplars classfication strategy and c) iCaRL-Hybrid1~\cite{rebuffi2017icarl}, also uses the exemplars but with a neural network classifier (i.e. a fully connected layer).  

\subsubsection{Using auxiliary exemplars or real exemplars}
we evaluate different methods which either using the auxiliary exemplars or real exemplars, shown in Table~\ref{tab:ablation_experiments_exemplar}. For a fair comparison, the memory cost for the auxiliary or the real is fixed. Our method is denoted by ``Ours.FC" if we utilize the fully connected layer as the classifier, or ``Ours.NCM" if utilizing the nearest-mean-of-exemplars classification strategy. Using auxiliary exemplars directly leads to a performance drop for both iCaRL and iCaRL-Hybrid1 because of the large domain gap between the auxiliary data and real data (shown in Figure~\ref{fig:tsne}(a)). For our domain-invariant learning method, the average accuracy of using the auxiliary exemplars is about $5\%$ higher than that of using the real exemplars. It seems to be proved that with the same memory buffer limitation, using auxiliary exemplars can further improve the performance compared with using the real exemplars, as long as the domain drift between them is reduced.


\begin{table}[t]
	\centering
	\caption{Validation of our duplet learning scheme and classifier adaptation scheme on iCIFAR-100 with the auxiliary samples of different fidelities. With our duplet learning scheme (DUP), the average accuracy of the class-incremental model is enhanced by more than 10\% compared to that of the normal learning scheme in~\cite{rebuffi2017icarl} for all cases.}
	\resizebox{1\columnwidth}{!}{
		\begin{threeparttable}
			\begin{tabular}{lllc}
				\toprule
				\multicolumn{2}{c}{Fidelity Factor $r$} &Method&Average Accuracy\\
				\midrule
				\midrule
				\multicolumn{2}{c}{\multirow{3}{*}{PCA $\frac{1}{3}$}}& iCaRL-Hybrid1&44.86$\pm$0.32\%\\
				
				&  &  iCaRL-Hybrid1+DUP&+14.67\%\\
				
				&  &  iCaRL-Hybrid1+DUP+CA&\textbf{+7.51\%}\\
				\midrule
				\multicolumn{2}{c}{\multirow{3}{*}{PCA $\frac{1}{6}$}}& iCaRL-Hybrid1&42.74$\pm$0.40\%\\
				&  & iCaRL-Hybrid1+DUP&+17.39\%\\
				&  & iCaRL-Hybrid1+DUP+CA&\textbf{+3.93\%}\\
				\midrule
				\multicolumn{2}{c}{\multirow{3}{*}{PCA $\frac{1}{12}$}}&iCaRL-Hybrid1&40.90$\pm$0.31\%\\
				&  & iCaRL-Hybrid1+DUP&+16.88\%\\
					&  & iCaRL-Hybrid1+DUP+CA&\textbf{+0.63\%}\\
				\midrule
				\multicolumn{2}{c}{\multirow{3}{*}{Downsampling $\frac{1}{3}$}}&iCaRL-Hybrid1&41.88$\pm$0.23\%\\
				& &iCaRL-Hybrid1+DUP&+16.41\%\\
					& &iCaRL-Hybrid1+DUP+CA&\textbf{+2.63\%}\\
				\midrule
				\multicolumn{2}{c}{\multirow{3}{*}{Downsampling $\frac{1}{6}$}}& iCaRL-Hybrid1&40.29$\pm$0.57\%\\
				&  & iCaRL-Hybrid1+DUP&+16.96\%\\
				&  & iCaRL-Hybrid1+DUP+CA&\textbf{+0.68\%}\\
				\bottomrule
			\end{tabular}
		\end{threeparttable}
	}
	
	\label{tab:ablation_experiments_DUP}%
\end{table}%

\begin{figure*}[t]
	\centering
	\subfigure{}{
		\begin{minipage}[ht]{0.31\textwidth}
			\includegraphics[width = 1\columnwidth]{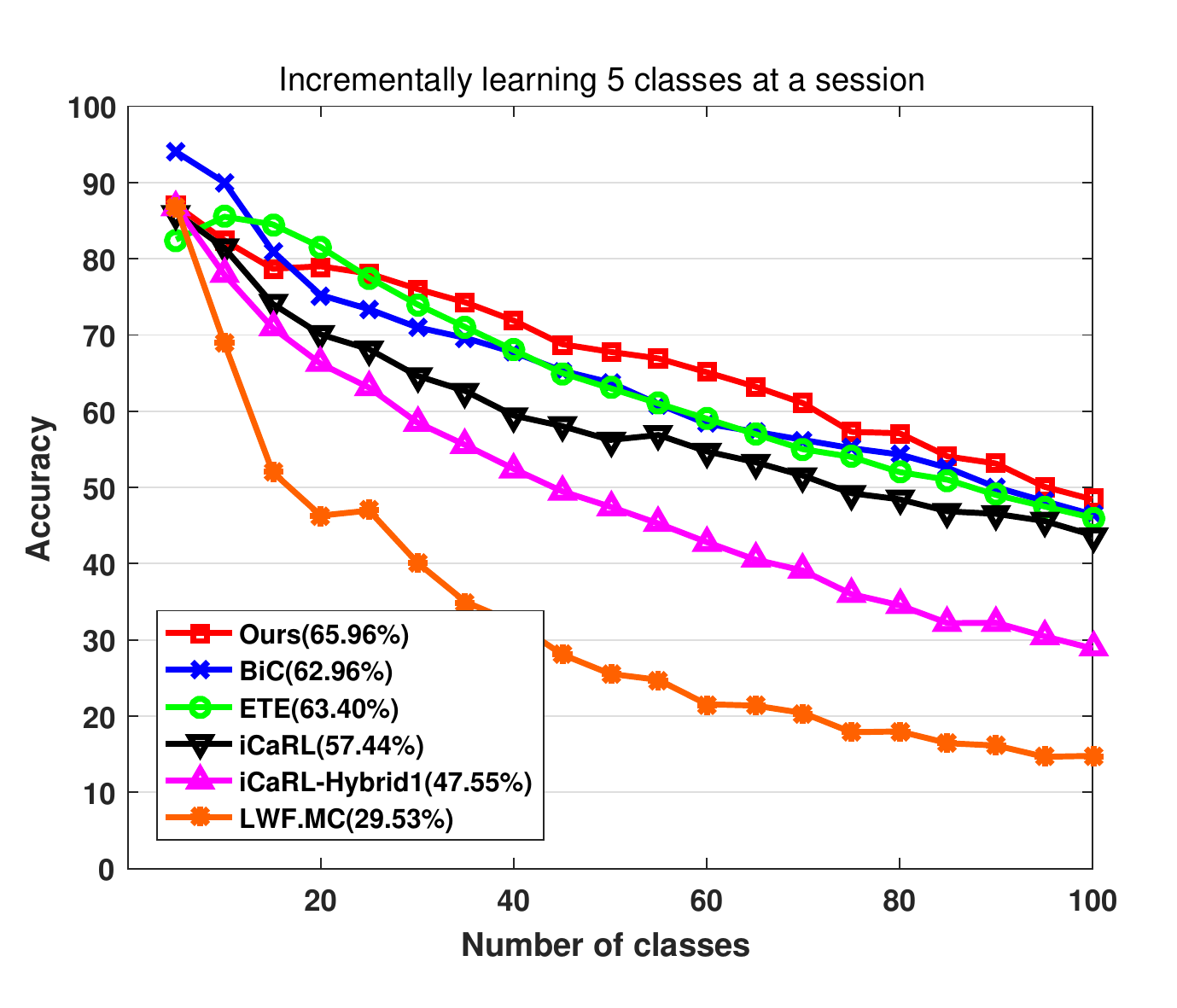}	
	\end{minipage}}
	\subfigure{}{
		\begin{minipage}[ht]{0.31\textwidth}
			\includegraphics[width = 1\columnwidth]{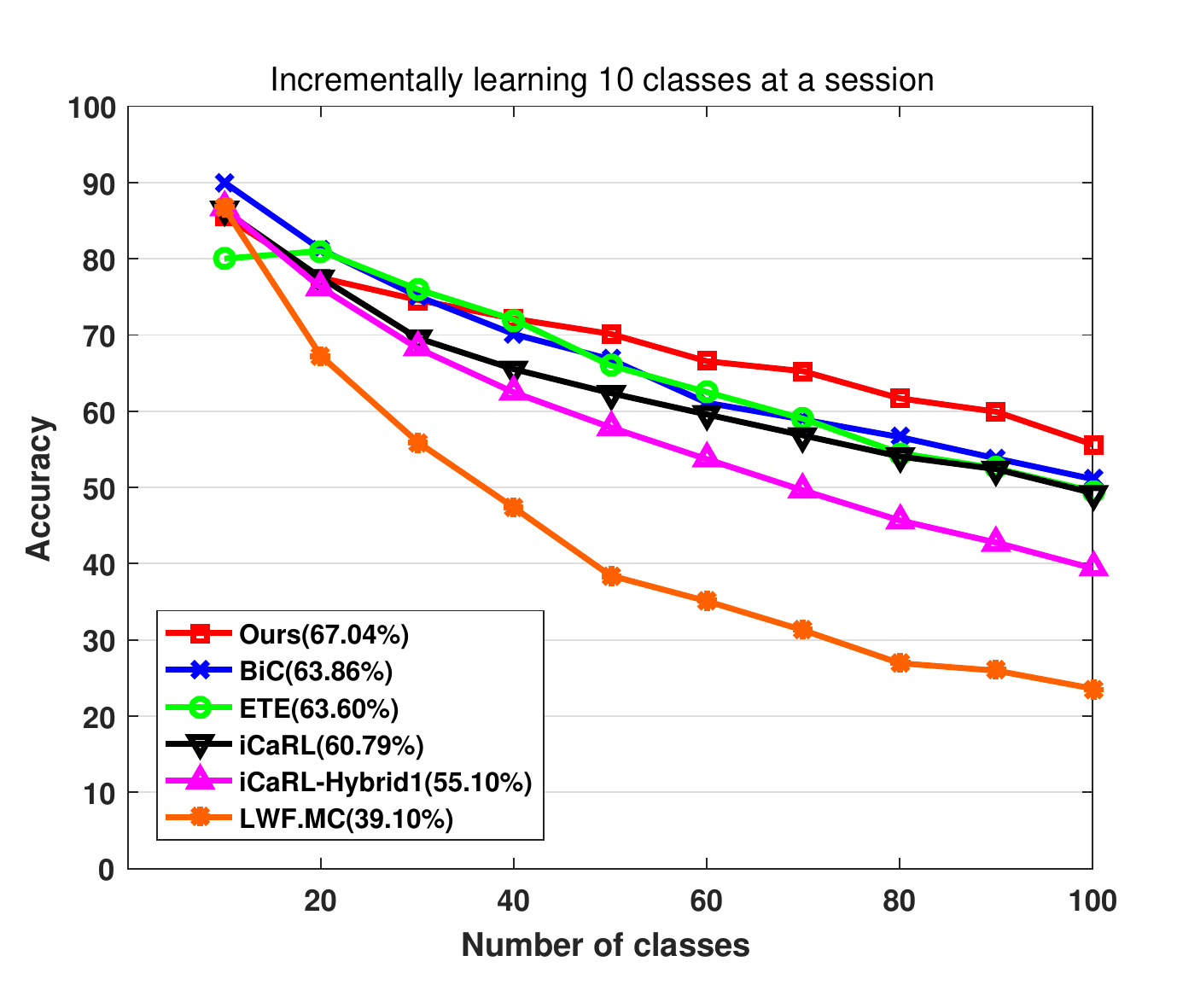}
		\end{minipage}	
	}
	\subfigure{}{
		\begin{minipage}[ht]{0.31\textwidth}
			\includegraphics[width = 1\columnwidth]{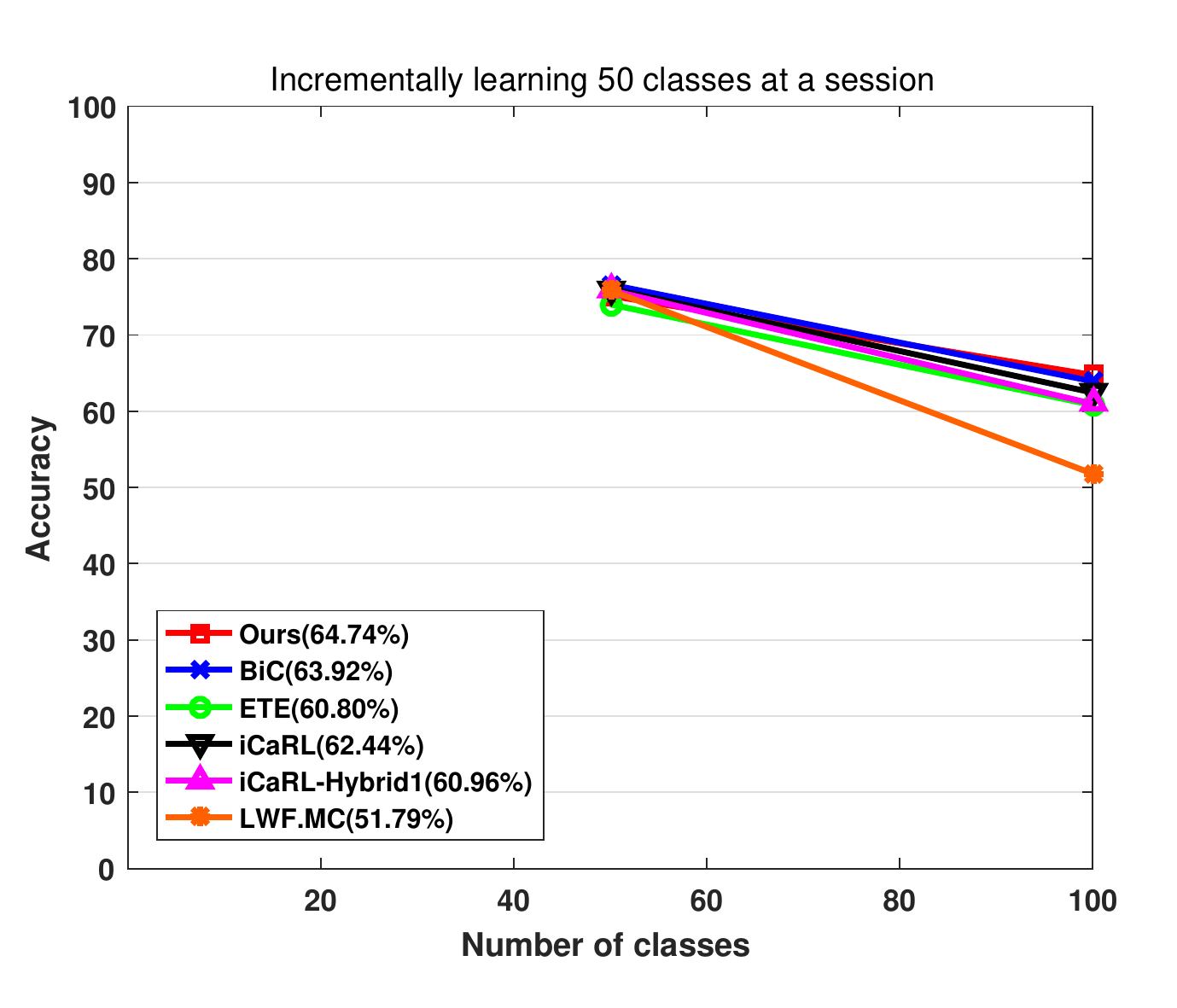}	
		\end{minipage}
	}
	\caption{The performance of different methods with the incremental learning session of 5, 10 and 50 classes on iCIFAR-100. The average accuracy of the incremental learning sessions is shown in parentheses for each method and computed without considering the accuracy of the first learning session. Our class-incremental learning scheme with auxiliary samples obtains the best results in all the cases.}
	\label{fig:state_of_art_cifar}
\end{figure*}
	
\begin{figure}[t]
	\centering
	\includegraphics[width=0.32\textwidth]{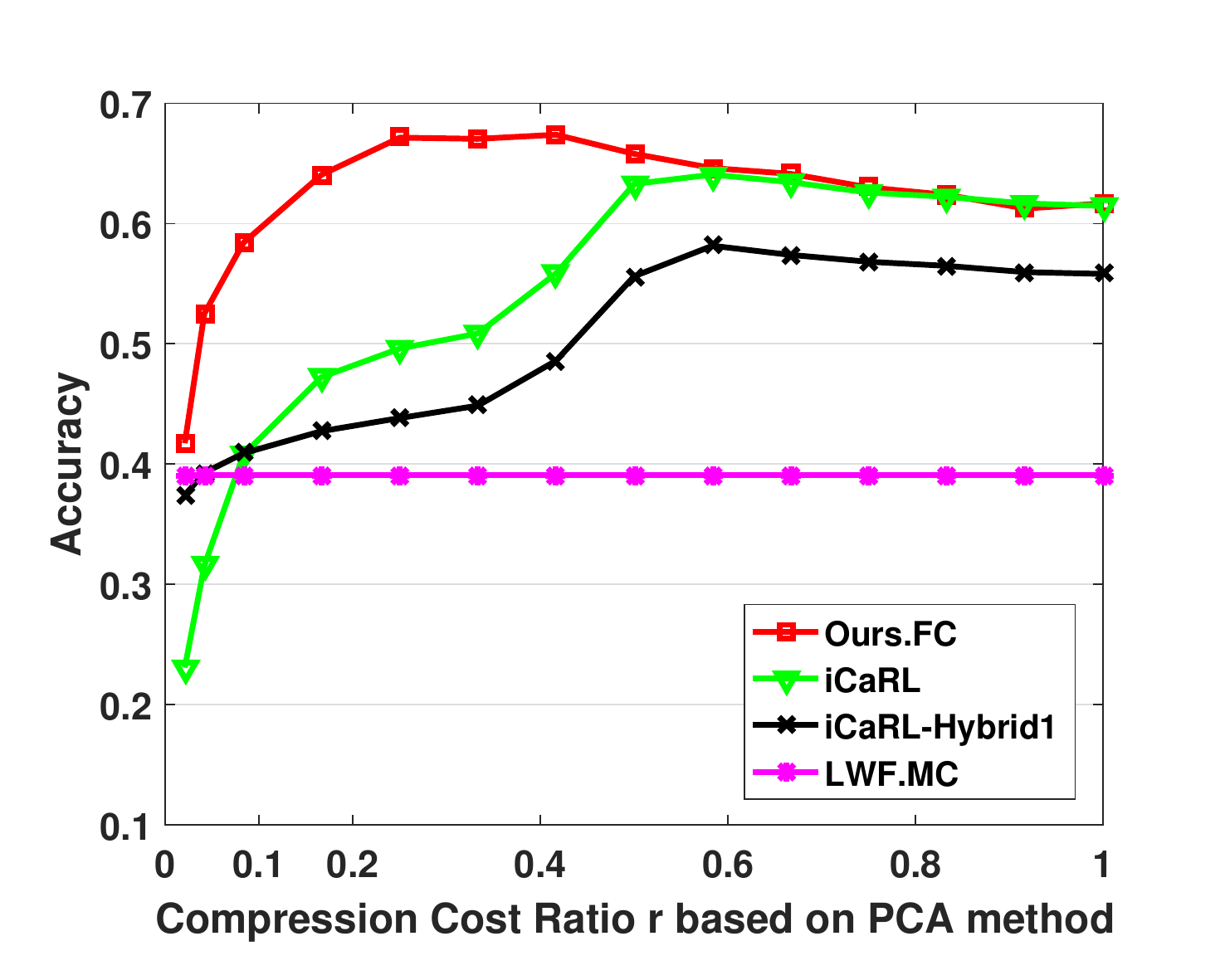}
	\caption{Performance of the model when varying the auxiliary sample's fidelity and the exemplar auxiliary data size. When $r=1$, it indicates that keeping the real exemplar samples in memory directly~\cite{rebuffi2017icarl}. }
	\label{fig:balance}
\end{figure}

\subsubsection{Validation of the duplet learning scheme}
we evaluate our duplet learning scheme with different auxiliary samples fidelity for the iCIFAR100 benchmark. $10$ new classes are added at each learning session. We utilize the ``iCaRL-Hybrid1" method with the auxilary samples by the normal learning scheme in~\cite{rebuffi2017icarl} or our duplet learning scheme (DUP). As shown in Table~\ref{tab:ablation_experiments_DUP}, we can observe that our scheme is able to significantly improve the performance of the final model greatly for various kinds of low-fidelity auxiliary samples compared with directly training the model. Moreover, the t-SNE~\cite{maaten2008visualizing} analysis in Figure~\ref{fig:tsne}(a) shows that normally there is a large gap between the auxiliary data and the real data without our duplet learning scheme. Figure~\ref{fig:tsne}(b) illustrates that our duplet learning scheme can actually reduce the domain drift and guarantee the effectiveness of the auxiliary data for preserving the model's performance on old classes.

\subsubsection{Validation of the classifier adaptation scheme}
we evaluate the performance of the final classifier after using our classifier adaptation scheme (CA). As shown in Table~\ref{tab:ablation_experiments_DUP}, the classifier adaptation scheme can further improve the classifier's accuracy. For the auxiliary samples of different fidelities based on the PCA or Downsampling, we can observe that the improvement of the performance decreases along with the auxiliary samples' fidelity. To investigate whether the CA scheme still works for other class-incremental learning methods, we have conducted experiments to evaluate our CA scheme with iCaRL-Hybrid1 and the results are shown in Table~\ref{tab:ablation_experiments_CA_2}. It can be observed that our scheme achieves more than $6.29\%$ average accuracy improvement over iCaRL-Hybrid1. For iCaRL-Hybrid1, there is also a classifier bias problem and the biased classifier can be refined by our classifier adaptation scheme, which leads to performance improvement.

\subsubsection{Balance of the  auxiliary exemplar samples' size and fidelity}
we examine the effect of varying the auxiliary sample's fidelity and the auxiliary exemplar data size for each class while the size of the limited memory buffer remains the same. Specifically, the fidelity will decrease if the size of auxiliary samples increase, where the number of the auxiliary samples is set to $\frac{2000}{r}$. As shown in Figure~\ref{fig:balance}, the average accuracy of the model increases first and then decreases with the decrease of the auxiliary samples' fidelity (also the increase of the auxiliary samples' size), which means moderately reducing the samples' fidelity can improve the final model's performance with a limited memory buffer. When the fidelity of an auxiliary sample is decreased a lot, the model's performance drops due to too much loss of class knowledge. 

\begin{figure}[t]
	\centering
	\includegraphics[width=0.32\textwidth]{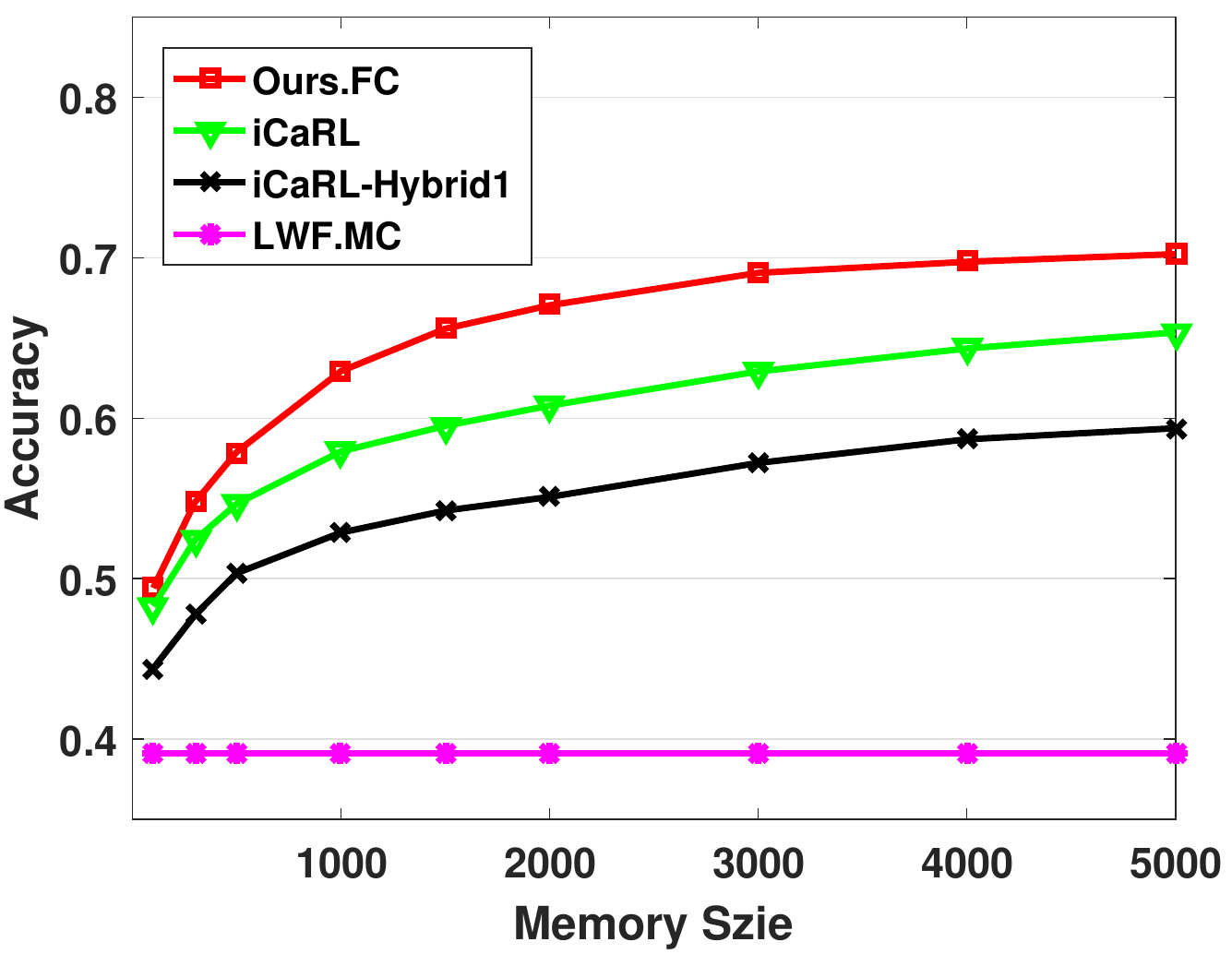}
	\caption{Average incremental accuracy on iCIFAR-100 with 10 classes per batch for the memory of different size (expressed in the number of real exemplar samples). The average accuracy of the model with our scheme is higher than that of iCaRL, LWF.MC and iCaRL-Hybrid1 in all the cases.}
	\label{fig:diff_budgets}		
\end{figure}

\begin{table}[t]
	\centering
	\caption{Evaluation of our classifier adaptation scheme for other class-incremental learning methods (e.g. iCaRL-Hybrid1) on iCIFAR-100.}
	\resizebox{1\columnwidth}{!}{
		\small
		\begin{threeparttable}
			\begin{tabular}{lllc}
				\toprule
				\multicolumn{2}{c}{Fidelity Factor $r$} &Method&Average Accuracy\\
				\midrule
				\midrule
				\multicolumn{2}{c}{\multirow{2}{*}{PCA $\frac{1}{3}$}}& iCaRL-Hybrid1&44.86$\pm$0.32\%\\
				
				&  &  iCaRL-Hybrid1+CA&\textbf{+9.40\%}\\
				\midrule
				\multicolumn{2}{c}{\multirow{2}{*}{PCA $\frac{1}{6}$}}& iCaRL-Hybrid1&42.74$\pm$0.40\%\\
				&  & iCaRL-Hybrid1+CA&\textbf{+9.17\%}\\
				\midrule
				\multicolumn{2}{c}{\multirow{2}{*}{PCA $\frac{1}{12}$}}&iCaRL-Hybrid1&40.90$\pm$0.31\%\\
				&  & iCaRL-Hybrid1+CA&\textbf{+6.29\%}\\
				\bottomrule
			\end{tabular}
		\end{threeparttable}
	}
	\label{tab:ablation_experiments_CA_2}%
\end{table}%

\begin{table*}[t]
	\centering
	\caption{Average incremental accuracy on iCIFAR-100 when changing the balance scalar $\lambda$.}
	\resizebox{0.9\textwidth}{!}{
		\begin{tabular}{cccccccccc}
			\toprule
			$\lambda$  &0&0.1&0.3&0.5& 0.8&1.0&2.0&5.0&10.0\\		
			\midrule
			Average Accuracy&30.55$\pm$0.47\%&50.89$\pm$0.17\%&64.57$\pm$0.25\%&66.43$\pm$0.15\%& 67.05$\pm$0.29\%&\textbf{67.11$\pm$0.26\%}&66.43$\pm$0.31\%&64.21$\pm$0.17\%&59.86$\pm$0.36\% \\
			\bottomrule						
		\end{tabular}
	}
	\label{tab:cifar:lam}
	\vspace{-10pt}
\end{table*}

\begin{table*}[t]
	\centering
	\caption{Average incremental accuracy on iCIFAR-100 with incremental laerning session of 10 classes when using a deep auto-encoder.}
	\resizebox{0.8\textwidth}{!}{
		\small
		\begin{tabular}{lcccccc}
			\toprule
			Memory Cost Ratio  &0.33\%&0.66\%&1.00\%&1.66\% &2.00\% &2.66\%\\
			\midrule
			Inter-sample Content Cost Ratio  &$\frac{20}{500}$&$\frac{20}{500}$&$\frac{20}{500}$&$\frac{20}{500}$&$\frac{20}{500}$&$\frac{20}{500}$ \\
			\midrule
			Compression Cost Ratio $r$&$\frac{1}{12}$&$\frac{1}{6}$&$\frac{1}{4}$&$\frac{5}{12}$&$\frac{1}{2}$&$\frac{2}{3}$\\
			\midrule		
			\midrule
			Ours-AE &52.13$\pm$0.31\%&56.95$\pm$0.38\%&59.29$\pm$0.46\%&61.21$\pm$0.32\%&\textbf{61.70$\pm$0.55\%}&60.82$\pm$0.26\%\\
			\bottomrule						
		\end{tabular}
	}
	\label{tab:cifar:redundancy_ratio}
	\vspace{-10pt}
\end{table*}
\begin{figure*}[t]
	\centering
	\subfigure[Session One]{
		\begin{minipage}[ht]{0.18\textwidth}
			\includegraphics[width = 1\columnwidth]{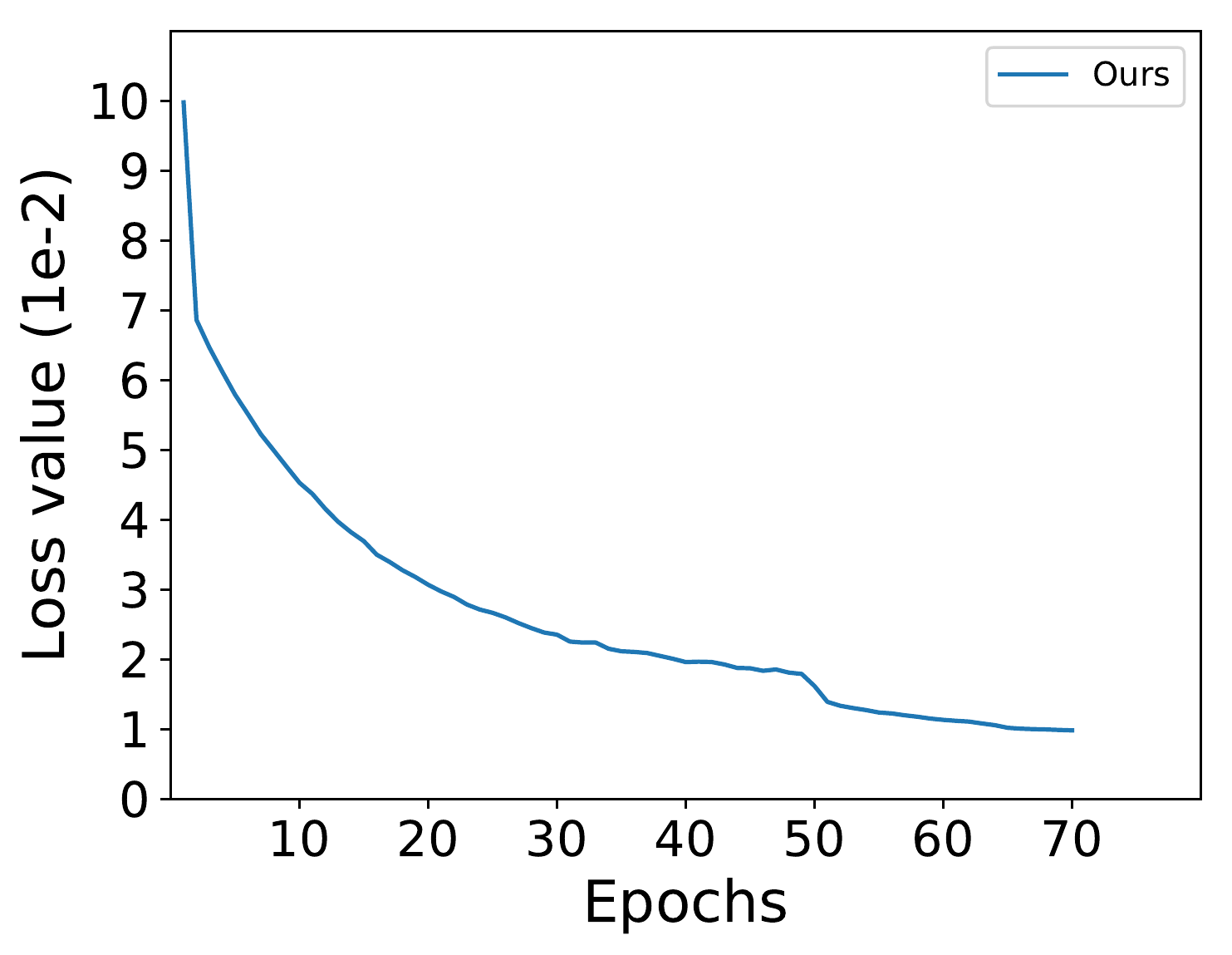}
		\end{minipage}	
	}	\subfigure[Session Two]{
		\begin{minipage}[ht]{0.18\textwidth}
			\includegraphics[width = 1\columnwidth]{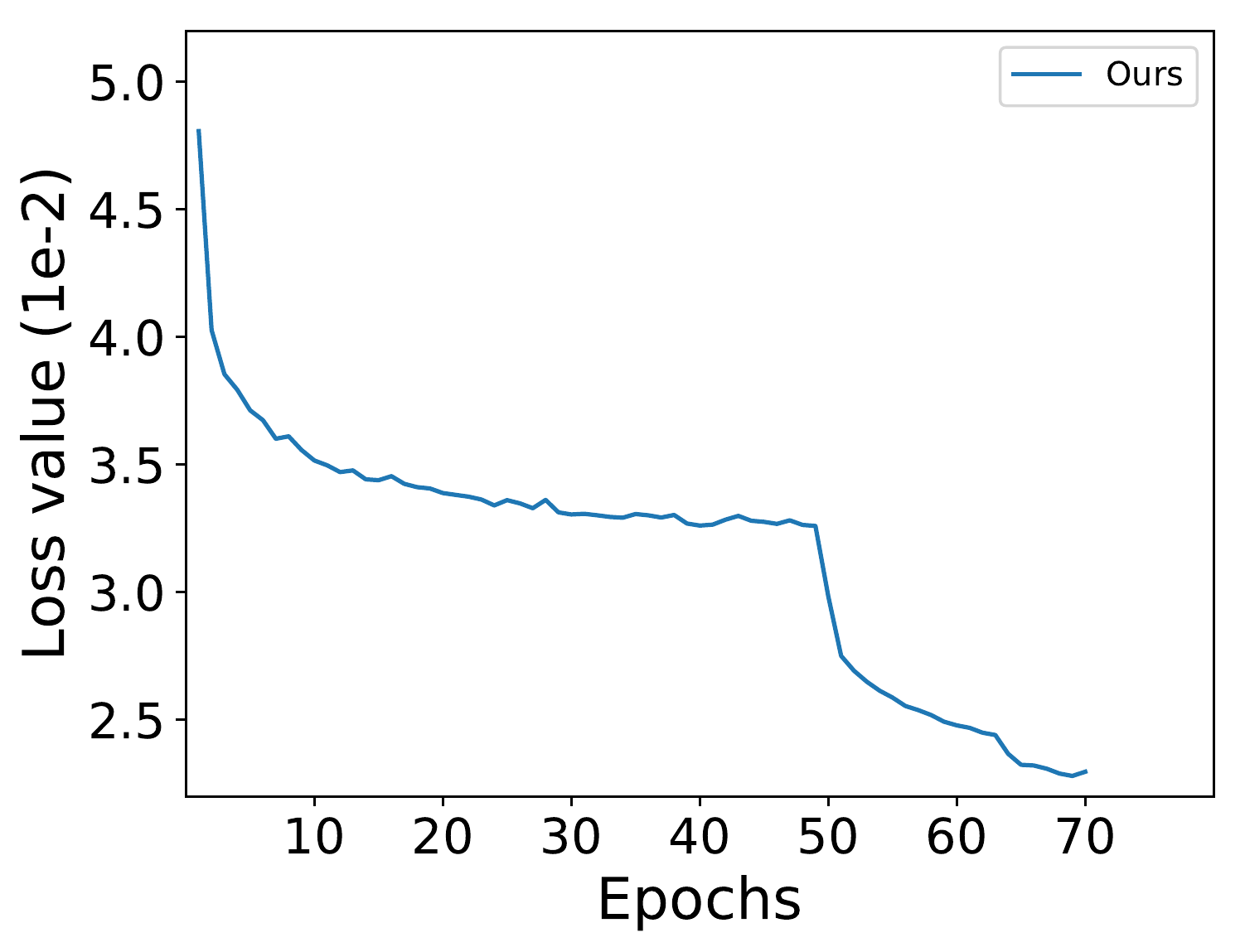}
		\end{minipage}	
	}
	\subfigure[Session Three]{
		\begin{minipage}[ht]{0.18\textwidth}
			\includegraphics[width = 1\columnwidth]{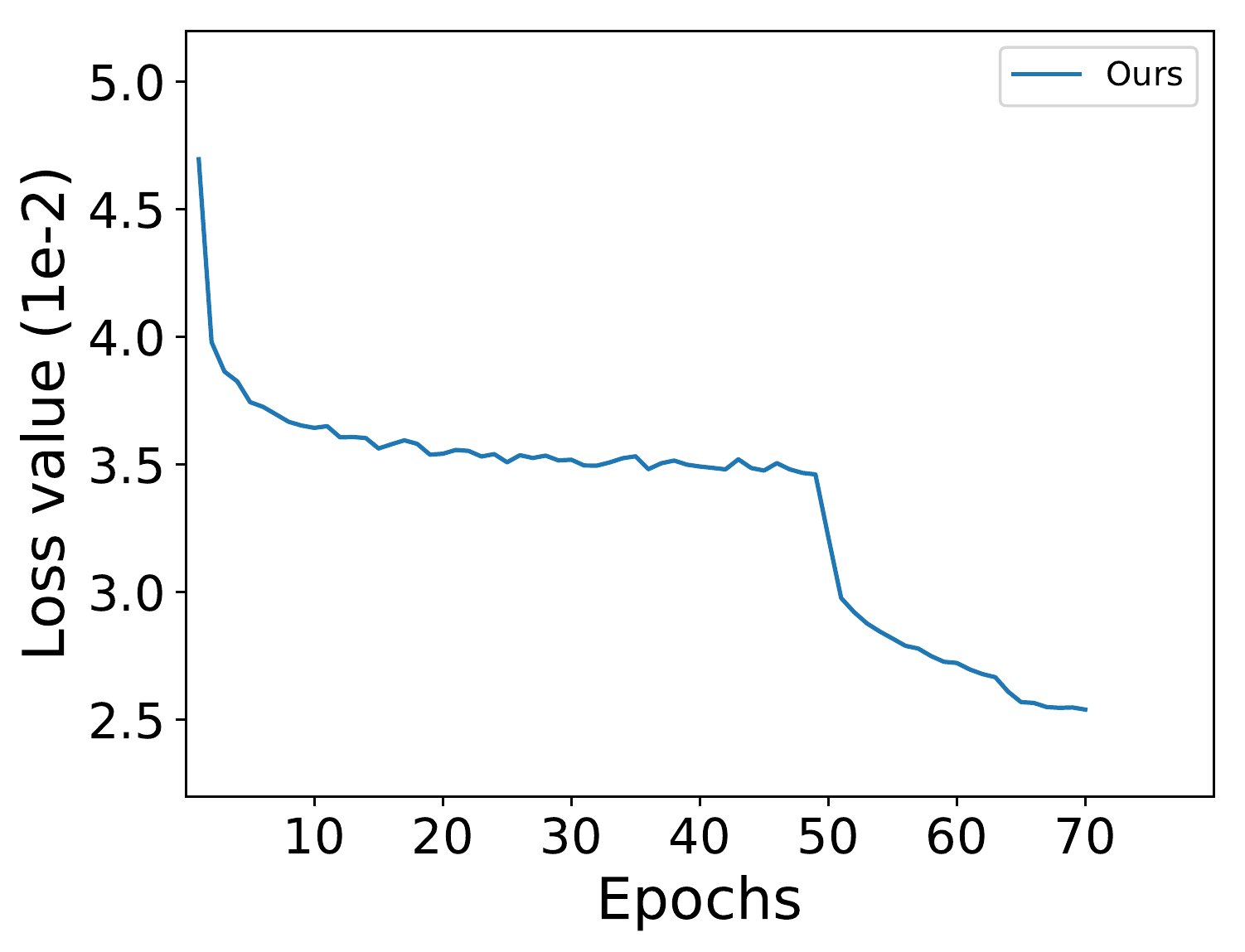}	
		\end{minipage}
	}
	\subfigure[Session Four]{
		\begin{minipage}[ht]{0.18\textwidth}
			\includegraphics[width = 1\columnwidth]{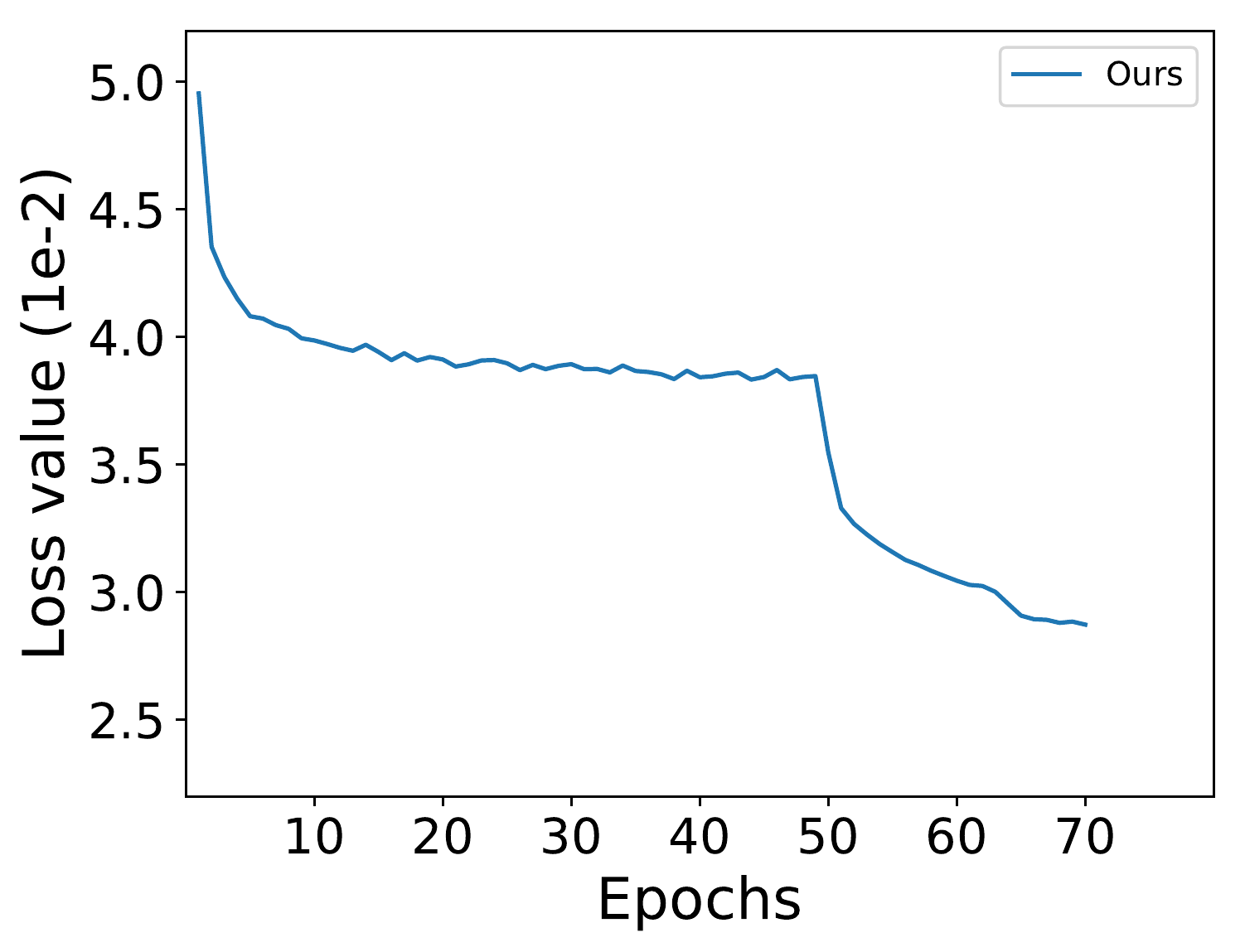}	
		\end{minipage}
	}
	\subfigure[Session Five]{
		\begin{minipage}[ht]{0.18\textwidth}
			\includegraphics[width = 1\columnwidth]{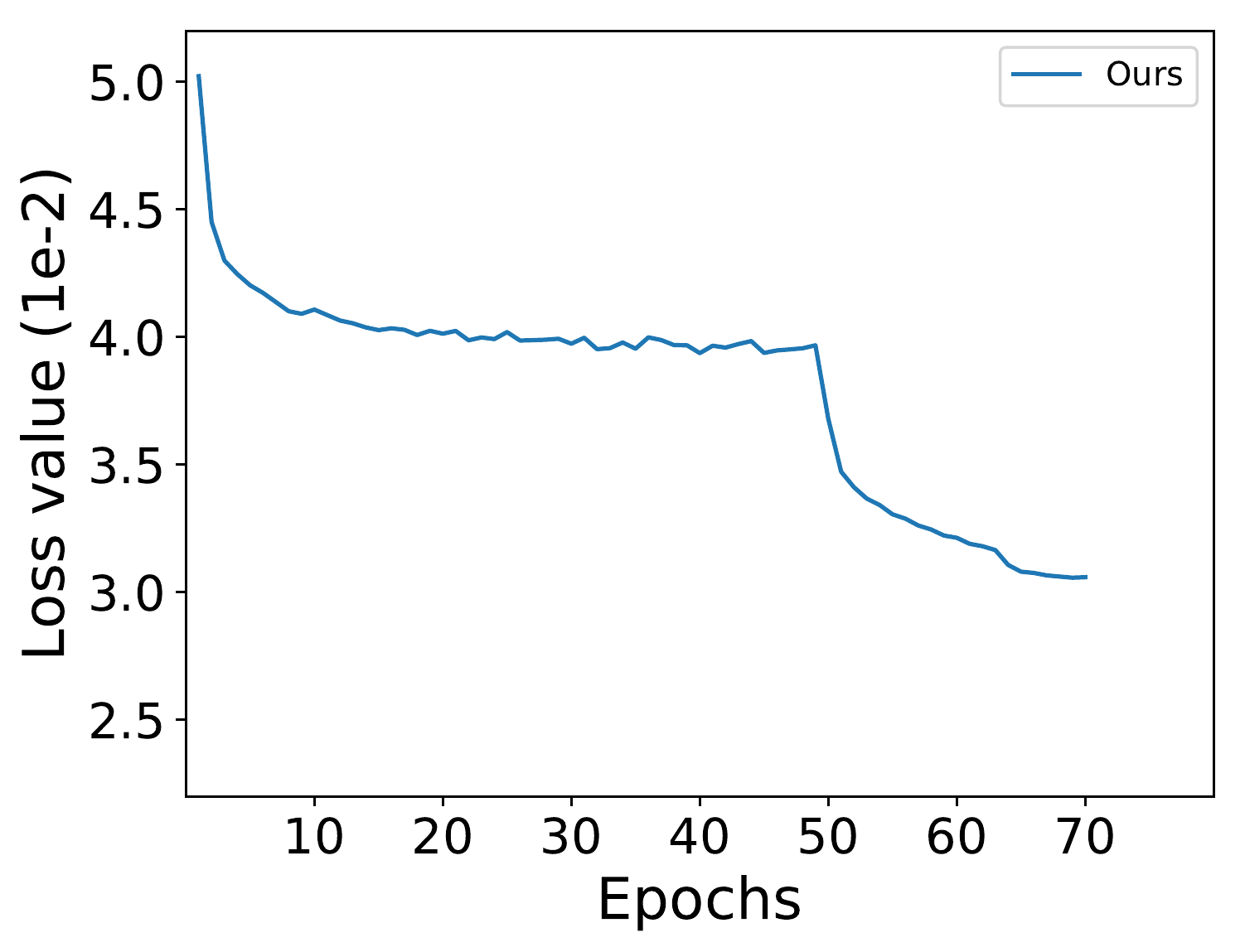}	
		\end{minipage}
	}
	\caption{Convergence analysis: the loss curve with the incremental learning session of 20 classes on iCIFAR100. (a) Session One, (b) Session Two, (c) Session Three, (d) Session Four, (e) Session Five.}
	\label{fig:loss_curve}
	\vspace{-10pt}
\end{figure*}
\begin{figure*}[t]
	\centering
	\subfigure[Session One]{
		\begin{minipage}[ht]{0.18\textwidth}
			\includegraphics[width = 1\columnwidth]{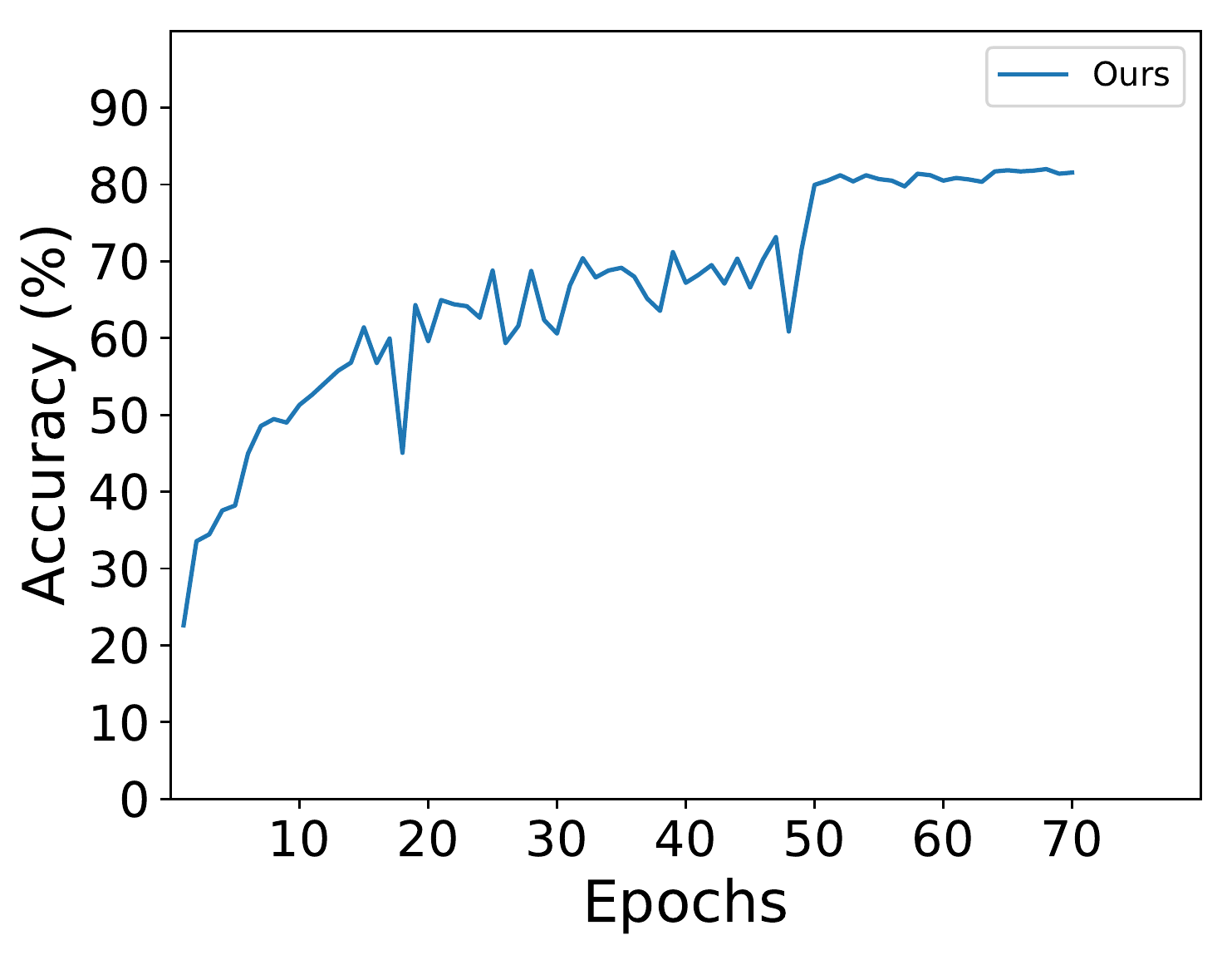}
		\end{minipage}	
	}	\subfigure[Session Two]{
		\begin{minipage}[ht]{0.18\textwidth}
			\includegraphics[width = 1\columnwidth]{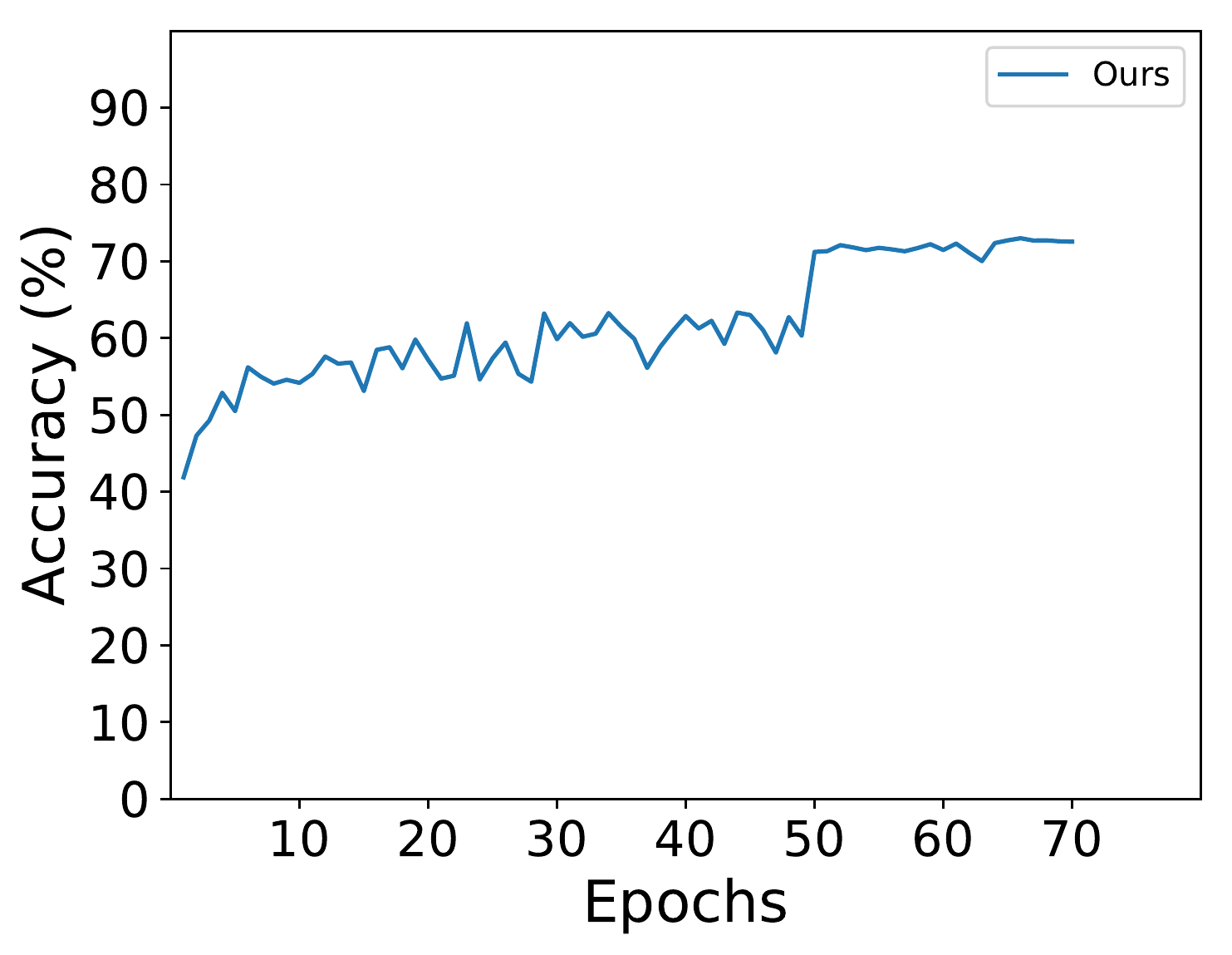}
		\end{minipage}	
	}
	\subfigure[Session Three]{
		\begin{minipage}[ht]{0.18\textwidth}
			\includegraphics[width = 1\columnwidth]{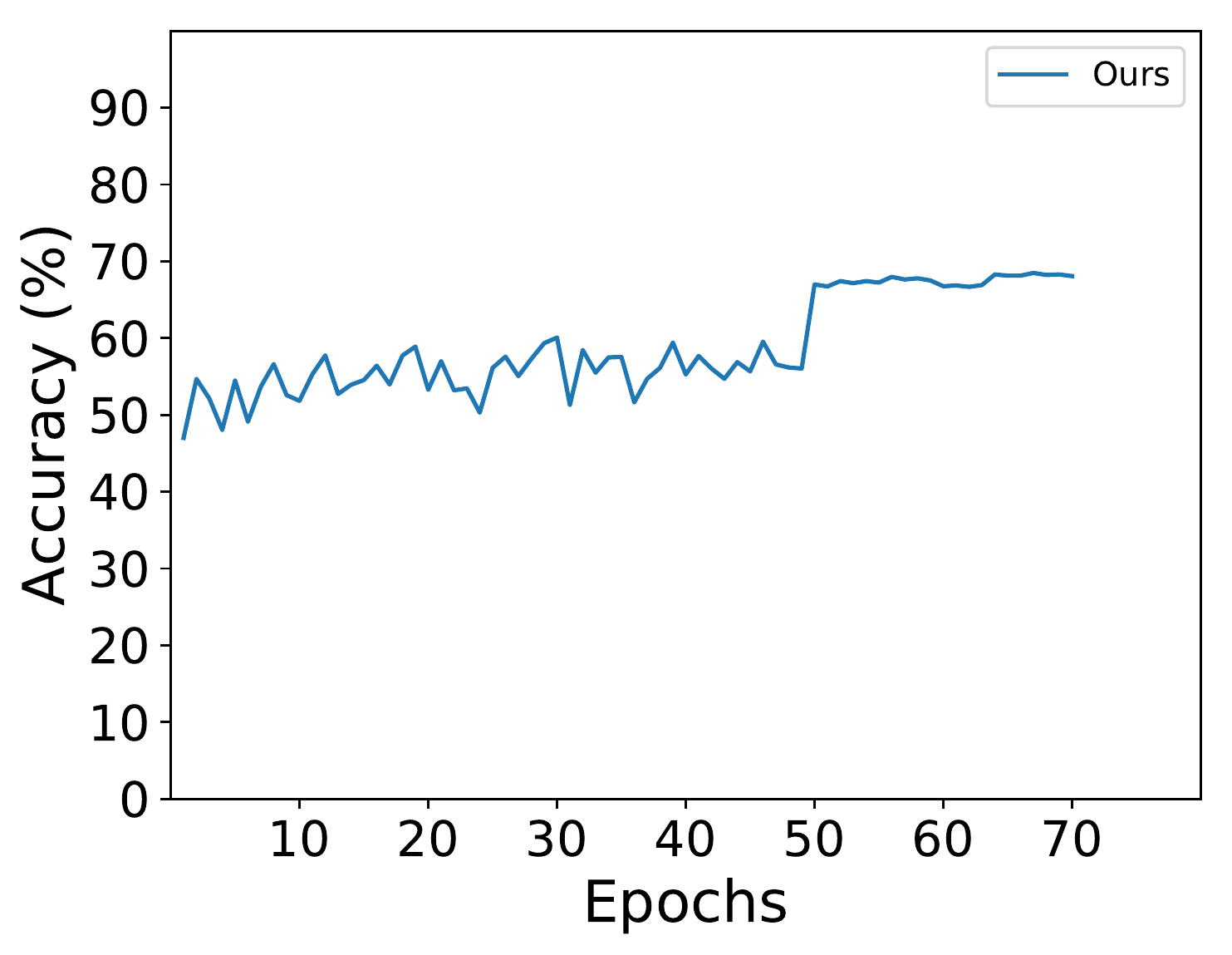}	
		\end{minipage}
	}
	\subfigure[Session Four]{
		\begin{minipage}[ht]{0.18\textwidth}
			\includegraphics[width = 1\columnwidth]{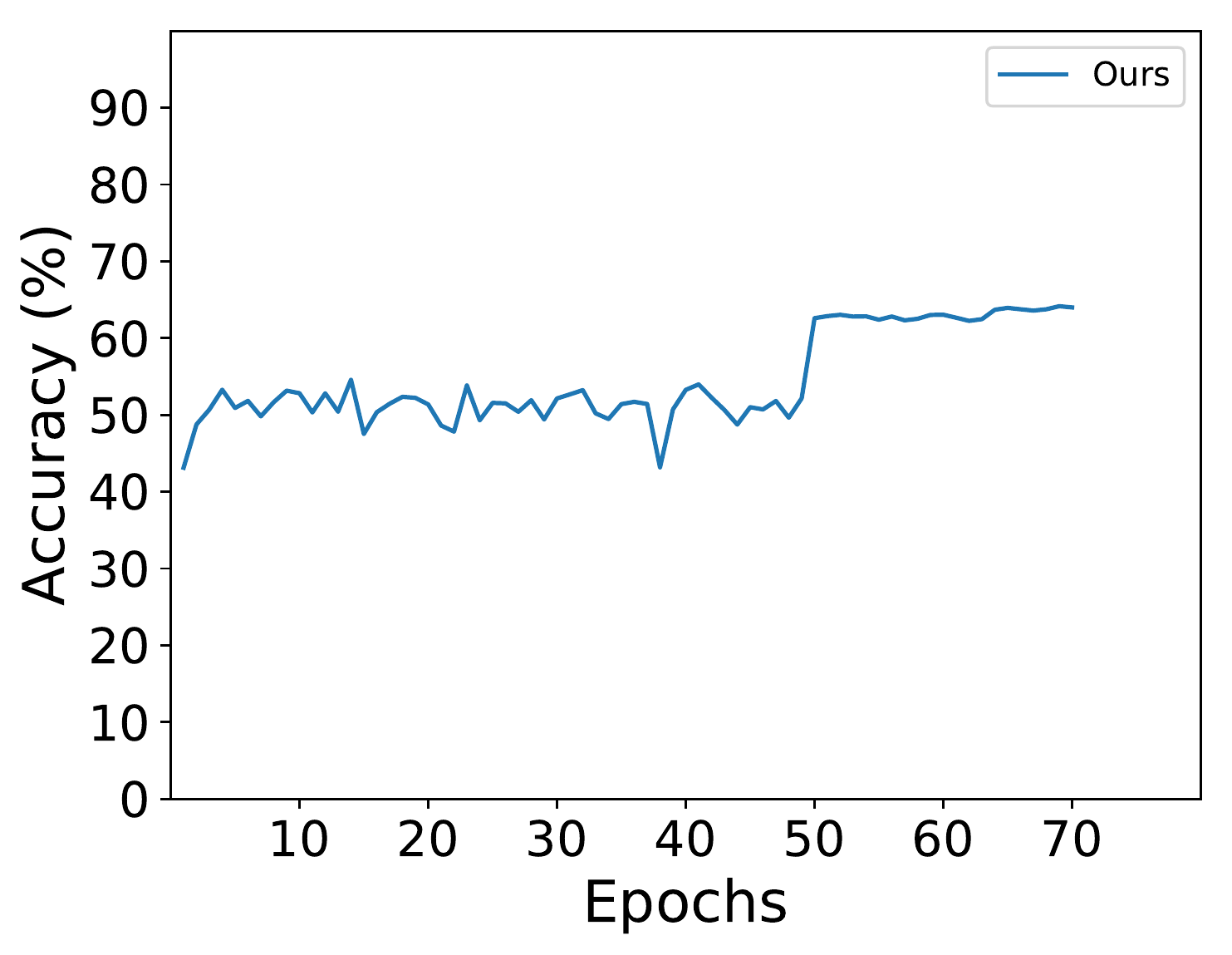}	
		\end{minipage}
	}
	\subfigure[Session Five]{
		\begin{minipage}[ht]{0.18\textwidth}
			\includegraphics[width = 1\columnwidth]{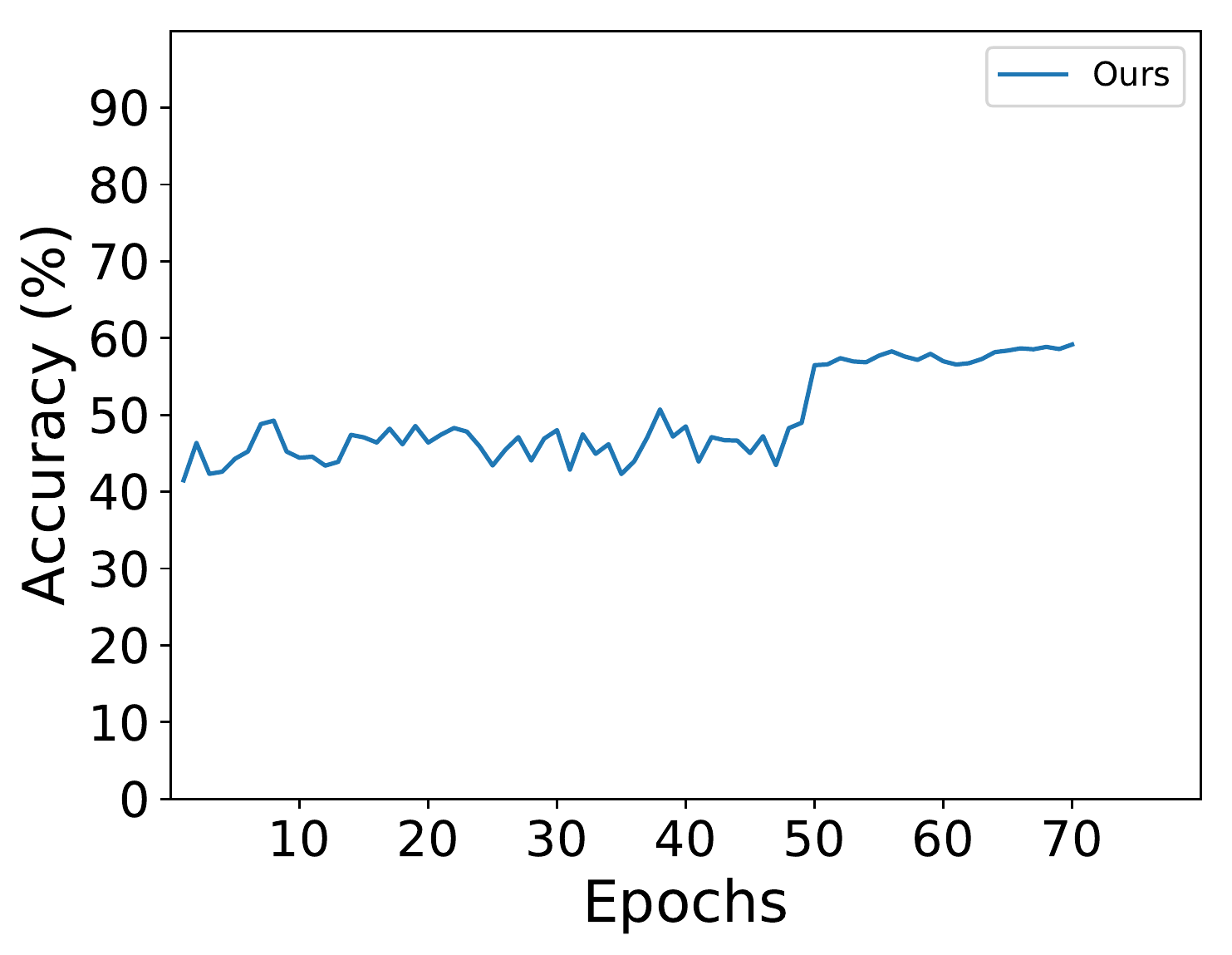}	
		\end{minipage}
	}
	\caption{Convergence analysis: the accuracy curve with the incremental learning session of 20 classes on iCIFAR100. (a) Session One, (b) Session Two, (c) Session Three, (d) Session Four, (e) Session Five.}
	\label{fig:accuracy}
	\vspace{-11pt}
\end{figure*}

\begin{table}[t]
	\centering
	\caption{Evaluation of our classifier adaptation scheme on iCIFAR-100 with the incremental learning session of 10 classes when varying the feature extractor.}
	\resizebox{1\columnwidth}{!}{
		\small
		\begin{threeparttable}
			\begin{tabular}{lllc}
				\toprule
				\multicolumn{2}{c}{Fidelity Factor $r$} &Method&Average Accuracy\\
				\midrule
				\midrule
				\multicolumn{2}{c}{\multirow{3}{*}{PCA $\frac{1}{3}$}}& iCaRL-Hybrid1+DUP&59.53$\pm$0.36\%\\
				&  &  iCaRL-Hybrid1+DUP+CA-varying&+6.06\%\\
				&  &  iCaRL-Hybrid1+DUP+CA-fixed&+7.51\%\\
				\bottomrule
			\end{tabular}
		\end{threeparttable}
	}
	
	\label{tab:cifar:varying}%
	\vspace{-10pt}
\end{table}%
\subsubsection{Fixed memory buffer size}
we conduct the experiments with memory buffer of different sizes on iCIFAR-100 where the number of added classes at each learning session is 10. The size of memory buffer is expressed in the number of real exemplar samples. As shown in Figure~\ref{fig:diff_budgets}, all of the exemplar-based methods (``Ours.FC", ``iCaRL" and ``iCaRL-Hybrid1") benefit from a larger memory which indicates that more samples of old classes are useful for keeping the performance of the model. The average accuracy of the model with our scheme is higher than that of iCaRL, LWF.MC and iCaRL-Hybrid1 in all the cases.

\begin{table}[t]
	\centering
	\caption{The average accuracy over all the incremental learning sessions except the first session on iCIFAR-100 with the incremental learning session of 20 classes.}
	\resizebox{0.35\textwidth}{!}{
		\small
		\begin{tabular}{lcc}
			\toprule
			Method  & Average Accuracy\\		
			\midrule
			LWF.MC~\cite{li2018learning}& 46.11\%\\
			iCaRL-Hybrid1~\cite{rebuffi2017icarl}& 61.18\%\\
			iCaRL~\cite{rebuffi2017icarl}& 63.33\%\\
			ETE~\cite{castro2018end}& 63.70\%\\
			BiC~\cite{Wu2019large}& 65.14\%\\
			ILCAN~\cite{xiang2019incremental}& 63.10\%\\
			ScaIL~\cite{belouadah2020scail}& 63.28\%\\
			\midrule
			Ours&\textbf{67.25$\pm$0.31\%}\\
			\bottomrule						
		\end{tabular}
	}
	\label{tab:cifar:state}
	\vspace{-11pt}
\end{table}

\subsubsection{Effect of the balance scalar $\lambda$}
To investigate the effect of the balance scalar $\lambda$, we have conducted experiments on iCIFAR-100 by changing $\lambda$ with the incremental learning session size being 10 classes. As shown in Table~\ref{tab:cifar:lam}, $\lambda=1.0$ leads to the best performance indeed under this setting and the performance of our method is stable within a relatively wide range from $0.5$ to $2.0$.

\subsubsection{Evaluation of CA scheme when varying the feature extractor}
we have conducted experiments to evaluate the CA scheme when varying the feature extractor on iCIFAR-100 with the incremental learning session of 10 classes. 
As shown in Table~\ref{tab:cifar:varying}, our CA scheme with the feature extractor fixed (CA-fixed) or varying (CA-varying) can further improve the performance by $6.06\%$ or $7.51\%$.	

\subsubsection{Evaluation of our method using a deep auto-encoder}
we have conducted experiments using a deep auto-encoder to compress each image on iCIFAR-100. In the experiments, we evaluated the performance of our method with different compression cost ratios by keeping the inter-sample content cost ratio fixed. The results are shown in Table~\ref{tab:cifar:redundancy_ratio}, where our method achieves the best performance when the compression cost ratio is $\frac{1}{2}$.

\begin{figure}[t]
	\centering
	\includegraphics[width=0.32\textwidth]{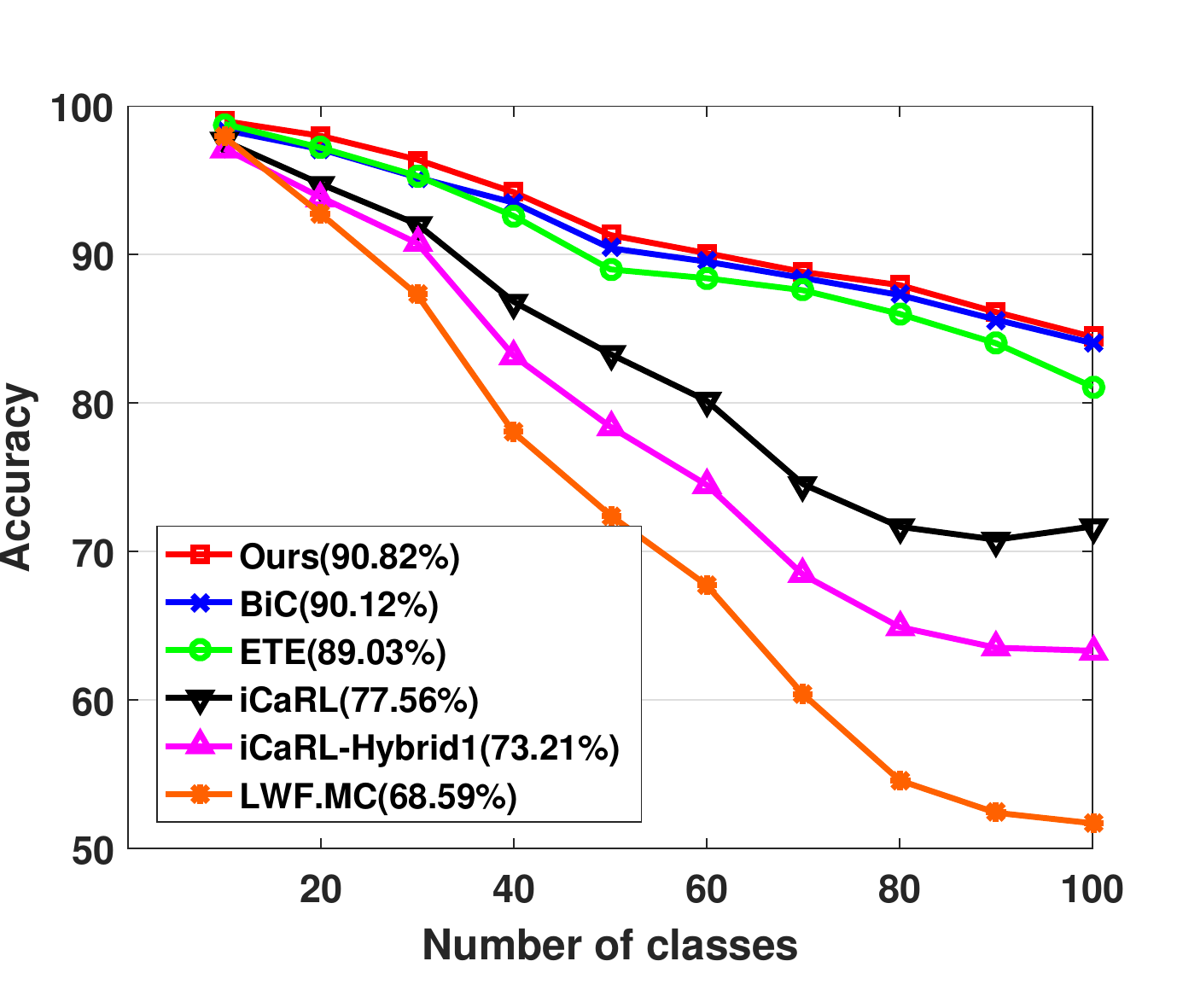}
	\caption{The performance of different methods with the incremental learning session of 10 classes on iILSVRC-small. The average accuracy over all the incremenal learning sessions is shown in parentheses for each method.}
	\label{fig:state_of_art_imagenet}
	\vspace{-11pt}
\end{figure}
%

\subsubsection{Convergence analysis of our method}
We evaluate the convergence performance with the incremental learning session of 20 classes on iCIFAR100. As shown in Figure~\ref{fig:loss_curve} and Figure~\ref{fig:accuracy}, we have shown the corresponding training loss curves and validation accuracy curves. From Figure~\ref{fig:loss_curve} and Figure~\ref{fig:accuracy}, we clearly observe that our method can achieve a relatively stable convergence performance for all the sessions.

\subsection{State-of-the-Art Performance Comparison}
In this section, we evaluate the performance of our proposed scheme on the iCIFAR-100 and iILSVRC benchmark, against the state-of-the-art methods, including LWF.MC~\cite{li2018learning}, iCaRL, iCaRL-Hybrid1~\cite{rebuffi2017icarl}, ETE~\cite{castro2018end}, ILCAN~\cite{xiang2019incremental}, ScaIL~\cite{belouadah2020scail}, BiC~\cite{Wu2019large}. 

For iCIFAR-100, we evaluate the incremental learning session of 5, 10, 20 and 50 classes. The memory size for all the evaluated methods is the same. As shown in Figure~\ref{fig:state_of_art_cifar}, we observe that our class-incremental learning scheme with auxiliary samples obtains the best results in all the cases. Compared with iCaRL and iCaRL-Hybrid1, we achieve a higher accuracy at each learning session. When the new-class data arrives, the accuracy of our scheme decreases slowly compared to ETE and BiC. As shown in Table~\ref{tab:cifar:state}, our method achieves better average accuracy than ILCAN and ScaIL.

For iILSVRC-small, we evaluate the performance of our method with the incremental learning session of 10 classes and the results are shown in Figure~\ref{fig:state_of_art_imagenet}. It can be also observed that 
our scheme obtains the highest accuracy at each incremental learning session among others. For iILSVRC-full, we evaluate the performance of our method with the incremental learning session of 100 classes and the results are shown in Table~\ref{tab:imagenet1000:state}. The average accuracy of our method over all the learning sessions except the first session is $2.1\%$ higher than BiC and outperforms the other methods by a large margin. Hence, our method performs consistently well on the datasets with a much larger scale of classes.  

\begin{table}[t]
	\centering
	\caption{The accuracy at the last incremental learning session and the average accuracy over all the incremental learning sessions except the first session on iILSVRC-full with the incremental learning session of 100 classes.}
	\resizebox{0.45\textwidth}{!}{
		\small
		\begin{tabular}{lcc}
			\toprule
			Method  &Last Accuracy & Average Accuracy\\		
			\midrule
			LWF.MC~\cite{li2018learning}&39.0\%&52.8\% \\
			iCaRL~\cite{rebuffi2017icarl}&44.0\%&60.8\% \\
			ETE~\cite{castro2018end}&52.0\%& 69.6\%\\
			BiC~\cite{Wu2019large}&73.2\%& 82.8\%\\
			\midrule
			Ours&\textbf{75.4$\pm$0.1\%}& \textbf{84.9$\pm$0.2\%}\\
			\bottomrule						
		\end{tabular}
	}
	\label{tab:imagenet1000:state}
	\vspace{-6pt}
\end{table}

\section{Conclusion}\label{conclusions}
In this paper, we have presented a novel memory-efficient exemplar preserving scheme and a duplet learning scheme for resource-constrained class-incremental learning, which transfers the old-class knowledge with low-fidelity auxiliary samples rather than the original real samples. We have also proposed a classifier adaptation scheme for the classifier's updating. The proposed scheme refines the biased classifier with samples of pure true class labels. Our scheme has obtained better results than the state-of-the-art methods on several datasets. As part of our future work, we plan to explore the low-fidelity auxiliary sample selection scheme for inheriting more class information in a limited memory buffer.

\section*{Acknowledgment}
This work is supported in part by National Key Research and Development Program of China under Grant 2020AAA0107400, National Natural Science Foundation of China under Grant U20A20222, Zhejiang Provincial Natural Science Foundation of China under Grant LR19F020004, and key scientific technological innovation research project by Ministry of Education.

\bibliographystyle{IEEEtran}
\bibliography{ref}{}

\end{document}